%% file: main.tex
\documentclass[10pt,twocolumn,letterpaper]{article}

\usepackage[pagenumbers]{cvpr}      
\usepackage{times}
\usepackage{latexsym}
\usepackage[T1]{fontenc}
\usepackage[utf8]{inputenc}
\usepackage{microtype}
\usepackage{inconsolata}
\usepackage{graphicx} 
\usepackage{booktabs}
\usepackage{multirow}
\usepackage{tabularx}
\usepackage{xspace}
\usepackage{pifont}
\usepackage{colortbl}
\newcommand{\crucible}{Crucible Bench\xspace}
\newcommand{\cmark}{\textcolor{green!55!black}{\ding{51}}}
\newcommand{\nomark}{\textcolor{black!30}{--}}
\definecolor{cvprblue}{rgb}{0.21,0.49,0.74}
\usepackage[pagebackref,breaklinks,colorlinks,allcolors=cvprblue]{hyperref}

\def\confName{CVPR}
\def\confYear{2027}

\title{Crucible: Evaluating LLM Agents in \\Executable Chemistry Research Environments}

\author{
Yinjun Liu$^{1*}$ \quad Xinye Li$^{2*}$ \quad Ruisen Peng$^{1*}$ \quad Lang Gao$^{1}$ \\
Zixiang Xu$^{3}$ \quad Zirui Song$^{1\dagger}$ \quad Xiuying Chen$^{1\ddagger}$ \\[0.5em]
$^1$Mohamed bin Zayed University of Artificial Intelligence \\
$^2$Northeastern University, Shenyang, China \quad $^3$University of Southern California \\[0.3em]
{\tt\small \{Yinjun.Liu, Zirui.Song, Xiuying.Chen\}@mbzuai.ac.ae} \quad {\tt\small lixinxinye@gmail.com} \\[0.3em]
{\small $^*$Equal contribution \quad $^\dagger$Project leader \quad $^\ddagger$Corresponding author}
}

\begin{document}
\maketitle

\begin{abstract}
Large language models (LLMs) perform well on chemistry knowledge benchmarks, which mainly assess question answering and isolated tasks. Chemical research, however, requires using specialized tools and interpreting their outputs to obtain chemically valid results. 
To address this gap, we introduce Crucible Bench to evaluate this ability across all four stages of the design–make–test–analyze (DMTA) cycle. 
The benchmark contains 1,078 tasks in 12 task families curated from 18 published datasets, covering molecular design, synthesis planning, experimental execution, closed-loop optimization, simulation, and analytical characterization. 
Agents access 75 domain-specific tools through the Model Context Protocol in seven isolated environments. Sandboxed graders verify the resulting synthesis routes, experimental protocols, and other outputs. Laboratory tasks use robotic simulators or previously measured experimental data. We evaluate 14 LLMs as agents and find that the best achieves an overall score of 0.545, with synthesis planning and closed-loop optimization emerging as the hardest domains. Overall performance does not predict optimization performance well: the model ranked twelfth overall achieves the highest score in closed-loop optimization. Error analysis also shows that agents often submit chemically invalid results despite running their tools successfully. Stronger models benefit more from tool access, but reliable use of chemical software remains a challenge that Crucible Bench can track as models improve.
\end{abstract}

\section{Introduction}

\suppressfloats[t]
\begin{figure}[t]
    \centering
    \includegraphics[width=\columnwidth]{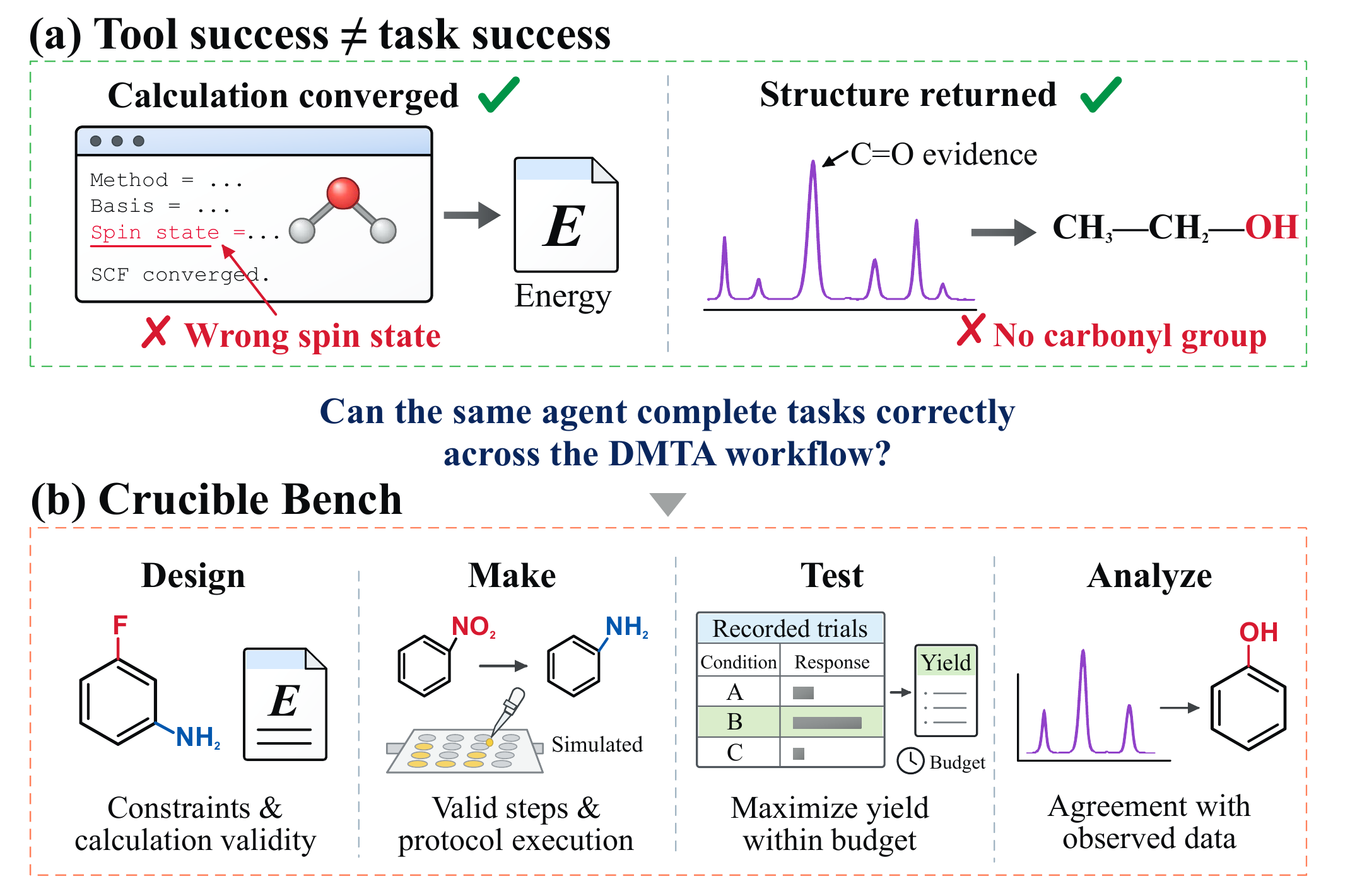}
    \caption{Successful tool execution does not guarantee scientifically valid results. \crucible evaluates agents across the DMTA research workflow by verifying their outputs against task-specific chemical, physical, and execution requirements.}
    \label{fig:overview}
\end{figure}

Large language models (LLMs) have made rapid progress on chemistry benchmarks. On curated question sets, the strongest models now outperform expert chemists on average \citep{mirza2025chembench}, and question-answering evaluations have documented their capabilities across a wide range of chemistry problems \citep{guo2023chemistry,rein2024gpqa,Chen_2025}. In parallel, agents that reason, write code, and invoke external tools \citep{yao2023react,schick2023toolformer,NEURIPS2025_fe9910d2} have been applied to synthesis planning \citep{bran2024chemcrow}, autonomous experimentation \citep{boiko2023coscientist}, and quantum-chemical workflows \citep{zou2025elagente,pham2026chemgraph}.
These advances raise a broader evaluation question: how reliably can agents use specialized tools to produce chemically valid results at different stages of research?

Evaluating this ability requires checking the artifacts that agents produce against the constraints of each task. For example, a density functional theory calculation with an incorrect spin multiplicity can still converge and return a plausible energy. A proposed synthesis route can conform to the expected format while combining several transformations into a single step. Similarly, a structural assignment must account for the measured spectra. Successful execution and well-formed outputs therefore provide limited evidence of chemical correctness. Evaluation shortcuts pose a further challenge, as tool-using agents can exploit weaknesses in scoring rules \citep{skalse2022defining,baker2025monitoring,zhong2026impossiblebench}. These examples motivate task-specific checks on the submitted artifacts.

The decisions needed to produce valid artifacts also vary across the research workflow. A typical project involves designing candidate molecules, computing their properties, planning and carrying out their synthesis, characterizing the products, and optimizing reaction conditions. Results from later stages can inform earlier decisions \citep{tom2024selfdriving}. Across these stages, a chemist may select a computational method, charge, and spin state using an electronic-structure package \citep{sun2020pyscf,bannwarth2019gfn2}, choose among routes proposed by a retrosynthesis planner \citep{genheden2020aizynthfinder}, translate a written procedure into commands for a liquid-handling system \citep{wierenga2023pylabrobot}, or test structural hypotheses with spectral and diffraction analysis tools \citep{ong2013pymatgen,fei2026dara}. Each task requires different decisions, so performance in one domain provides only a partial picture of an agent's capabilities.

Existing evaluations leave much of the chemistry research workflow untested. Execution-based benchmarks assess agents through the outcomes of their actions in software engineering and interactive environments \citep{jimenez2024swebench,liu2024agentbench,yao2025taubench}, while related scientific evaluations address data-driven discovery \citep{chen2025scienceagentbench}, biology research \citep{laurent2024labbench}, and laboratory safety \citep{zhou2026labsafety}. Chemistry benchmarks typically focus on specific problems, such as scenario-grounded molecular design \citep{jeong2026moldesignbench}, procurement cost estimation \citep{wu2026chemcost}, or robotic workflow planning for catalysis \citep{guo2026stress}. This narrow coverage makes it difficult to determine whether the same agent can use specialized tools and produce chemically valid results across different stages of research. We introduce \crucible to evaluate this ability across all four stages of the design--make--test--analyze (DMTA) cycle, as illustrated in Figure~\ref{fig:overview}. The benchmark comprises 1,078 tasks curated from 18 published datasets and organized into 12 task families across six domains: molecular design, synthesis planning, experimental execution, closed-loop optimization, simulation, and analytical characterization. Agents access 75 domain-specific tools through the Model Context Protocol (MCP) across seven isolated environments. For each task, they submit an artifact, such as a synthesis route, a computed energy, a structured experimental protocol, or a phase assignment, which a sandboxed grader checks against task-specific requirements. Experimental execution tasks use robotic liquid-handling simulation, and closed-loop optimization tasks return published measurements for the conditions selected by the agent.

We make three contributions. Firstly, we introduce \crucible, an execution-based benchmark that evaluates chemistry agents across the full design--make--test--analyze cycle. Secondly, we provide isolated environments with domain-specific tools and sandboxed graders that verify submitted artifacts against task-specific requirements. Finally, we evaluate LLM agents across the workflow and identify synthesis planning and closed-loop optimization as the most challenging domains. Our error analysis shows that agents often submit chemically invalid results even when their tools run successfully.

 \begin{figure*}[t]
    \centering
    \includegraphics[width=\textwidth]{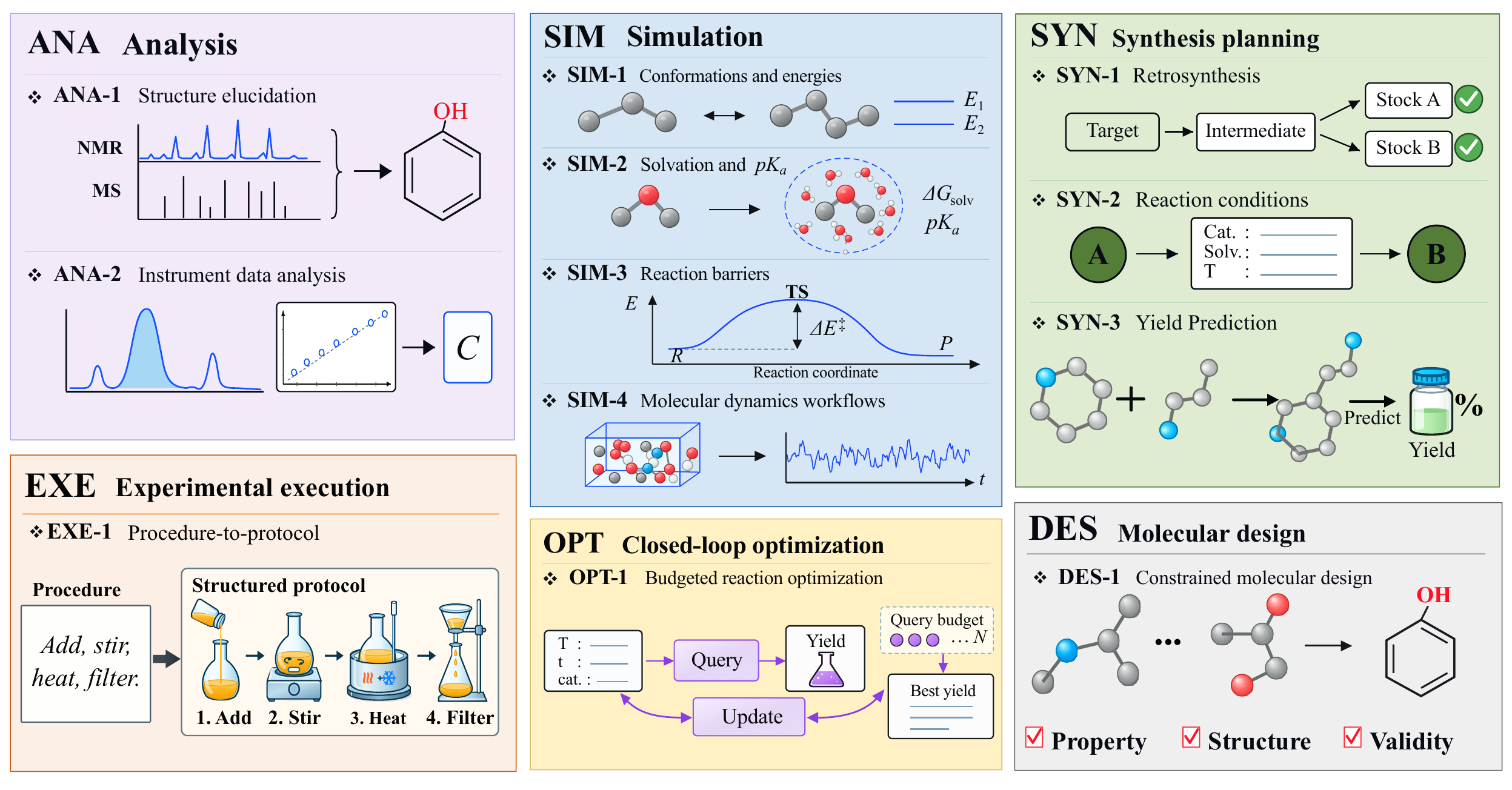}
    \caption{\textbf{Task families and representative outputs in \crucible.}
    The benchmark covers 12 task families across six domains: molecular design (DES), simulation (SIM), synthesis planning (SYN), experimental execution (EXE), analytical characterization (ANA), and closed-loop optimization (OPT).
    Tasks are evaluated independently using task-specific grading criteria.
    All illustrations are schematic.}
    \label{fig:task_families}
\end{figure*}

\section{Related Work}

\subsection{LLMs for Chemistry}

Large language models have been adapted to chemistry through domain-specific training~\citep{song-etal-2025-injecting}.
Mol-Instructions~\citep{fang2024molinstructions} releases a large instruction dataset covering molecules, proteins and biomolecular text, LlaSMol~\citep{yu2024llasmol} fine-tunes open models on a broad set of small-molecule tasks such as name conversion, property prediction and reaction prediction, and ChemLLM~\citep{zhang2024chemllm} builds a dialogue model from structured chemical data converted into instructions.
ether0~\citep{narayanan2025ether0} instead uses reinforcement learning to train a reasoning model that answers chemistry questions with molecular structures.

\crucible evaluates such LLMs as the reasoning core of an agent, testing whether a model can complete chemistry tasks correctly using domain-specific tools.

\subsection{Chemistry Agents}

An LLM becomes a chemistry agent when it is given chemistry tools and a control loop in which it alternates between reasoning and tool calls~\citep{yao2023react}.
ChemCrow~\citep{bran2024chemcrow} pioneered this design by connecting GPT-4 to 18 expert-designed tools for organic synthesis, drug discovery and materials design, and Coscientist~\citep{boiko2023coscientist} combined an LLM with web search, code execution and laboratory automation to plan and run experiments.
Later systems target individual stages of chemical research, including end-to-end synthesis development~\citep{ruan2024llmrdf}, robotic experimentation with multiple cooperating agents~\citep{song2025chemagents}, quantum-chemical calculations~\citep{zou2025elagente}, molecular dynamics~\citep{campbell2026mdcrow} and electrocatalyst discovery~\citep{zhang2025crest}.
Other work studies the agents themselves, such as self-updating memory~\citep{tang2025chemagent} and the effect of tool use on chemical problem solving~\citep{yu2025tooling}.
Each system, however, builds its own tools for its own tasks and ties them to its own framework, so the tools cannot be reused or compared across systems.

\crucible instead packages the software used across the whole workflow, from electronic-structure codes and retrosynthesis planners to liquid-handling simulators and diffraction analysis, as 75 tools on seven isolated MCP servers that any model or agent framework can call.

\subsection{Benchmarks for Scientific Agents}

Chemical capabilities of LLMs are most often measured with question answering.
\citet{guo2023chemistry} evaluate eight practical chemistry tasks, from name prediction to reaction and yield prediction, and ChemBench~\citep{mirza2025chembench} assembles more than 2,700 questions on which the strongest models outperform expert chemists on average, and ScholarChemQA~\citep{Chen_2025} builds multiple-choice questions from chemistry research papers.
GPQA~\citep{rein2024gpqa} and SciKnowEval~\citep{feng2025sciknoweval} pose graduate-level and multi-level scientific questions that include chemistry, while ChemIQ~\citep{runcie2026chemiq} and ChemCoTBench~\citep{li2025chemcotbench} move from recall to reasoning, asking models to interpret spectra or to edit and transform molecules step by step.

Benchmarks for scientific agents instead require models to act.
ScienceAgentBench~\citep{chen2025scienceagentbench} and DiscoveryBench~\citep{majumder2025discoverybench} evaluate data-driven discovery through program generation, LAB-Bench~\citep{laurent2024labbench} and BioPlanner~\citep{odonoghue2023bioplanner} target biology research and protocol planning, SciAgentGym~\citep{shen2026sciagentgym} and FrontierChallenge~\citep{su2026frontierchallenge} evaluate multi-step tool use and workflow completion across disciplines, and \citet{riosgarcia2026corral} build executable environments to analyse how agents reason in chemistry and materials science.
Within chemistry, executable benchmarks target individual stages, such as molecular dynamics~\citep{anand2026mdarena,kumar2026mdgym} and molecular design~\citep{han2026smddbench,jeong2026moldesignbench}, or individual tasks such as procurement cost estimation~\citep{wu2026chemcost}, laboratory safety~\citep{zhou2026labsafety} and robotic workflow planning for catalysis~\citep{guo2026stress}.
As Table~\ref{tab:comparison} shows, general-purpose scientific benchmarks cover few chemistry domains, and chemistry benchmarks each cover a single stage.

\crucible is, to our knowledge, the first to evaluate chemistry agents across all four stages of the design--make--test--analyze cycle, in the software chemists use, verifying their artifacts against published references.

\begin{table}[t]
\caption{Coverage of the six task domains of \crucible{} by executable
benchmarks for scientific and chemistry agents, grouped by DMTA stage: design
(D), make (M), test (T) and analyze (A).}
\label{tab:comparison}
\centering
\small
\setlength{\tabcolsep}{1pt}
\begin{tabularx}{\columnwidth}{@{}lr*{6}{>{\centering\arraybackslash}X}@{}}
\toprule
 & & D & \multicolumn{3}{c}{M} & T & A \\
\cmidrule(lr){3-3}\cmidrule(lr){4-6}\cmidrule(lr){7-7}\cmidrule(l){8-8}
Benchmark & Tasks & DES & SYN & EXE & OPT & SIM & ANA \\
\midrule
ScienceAgentBench~\citep{chen2025scienceagentbench} & 102 & \nomark & \nomark & \nomark & \nomark & \nomark & \nomark \\
SciAgentGym~\citep{shen2026sciagentgym} & 259 & \nomark & \nomark & \nomark & \nomark & \nomark & \nomark \\
FrontierChallenge~\citep{su2026frontierchallenge} & 97 & \nomark & \nomark & \nomark & \nomark & \cmark & \cmark \\
\midrule
MDArena~\citep{anand2026mdarena} & 50 & \nomark & \nomark & \nomark & \nomark & \cmark & \nomark \\
MDGym~\citep{kumar2026mdgym} & 169 & \nomark & \nomark & \nomark & \nomark & \cmark & \nomark \\
SMDD-Bench~\citep{han2026smddbench} & 502 & \cmark & \nomark & \nomark & \nomark & \nomark & \nomark \\
MolDesignBench~\citep{jeong2026moldesignbench} & 2,000 & \cmark & \nomark & \nomark & \nomark & \nomark & \nomark \\
ChemCost~\citep{wu2026chemcost} & 1,427 & \nomark & \nomark & \nomark & \nomark & \nomark & \nomark \\
Guo et al.~\citep{guo2026stress} & 32 & \nomark & \nomark & \cmark & \nomark & \nomark & \nomark \\
\midrule
\rowcolor{gray!12}
\textbf{\crucible} & 1,078 & \cmark & \cmark & \cmark & \cmark & \cmark & \cmark \\
\bottomrule
\end{tabularx}
\end{table}

\section{\crucible}
\label{sec:benchmark}
\crucible evaluates whether LLM agents can effectively use chemical research software to produce chemically valid results. It covers the full design--make--test--analyze (DMTA) cycle with tasks that reflect the work of each stage (\S~\ref{sec:construction}). Agents use established chemical software to carry out these tasks, configuring the tools and interpreting their outputs (\S~\ref{sec:environments}). Task-specific verifiers check the submitted artifacts for chemical correctness (\S~\ref{sec:verification}). The benchmark contains 1,078 test tasks in 12 task families, illustrated in Figure~\ref{fig:task_families} and specified in Table~\ref{tab:families}.

\begin{table*}[t]
\caption{\textbf{Task families and scoring criteria in \crucible.}
Tasks are grouped by design, make, test, and analyze stages. DES denotes molecular design, SYN synthesis planning, EXE experimental execution, OPT closed-loop optimization, SIM simulation, and ANA analytical characterization. For each task family, we report the required submission, scoring criteria, and number of test tasks.}
\label{tab:families}
\centering
\small
\setlength{\tabcolsep}{5pt}
\begin{tabularx}{\textwidth}{lll >{\raggedright\arraybackslash}X >{\raggedright\arraybackslash}X r}
\toprule
Stage & Domain & Family & Artifact & Grading & Tasks \\
\midrule
Design & DES & DES-1 Constrained design & Molecule satisfying constraints, or \texttt{null} if infeasible & All source checks pass & 163 \\
\midrule
\multirow{5}{*}{Make} & \multirow{3}{*}{SYN} & SYN-1 Retrosynthesis & Up to 10 multi-step routes to purchasable stock & Reference route in top 10 & 100 \\
 & & SYN-2 Conditions & Up to 10 sets of reagents, solvents and temperature & All roles match, $\pm$10\,\textdegree C & 150 \\
 & & SYN-3 Yield prediction & Measured yield (\%) of a reaction run under given conditions & Within 10 percentage points & 50 \\
\cmidrule{2-6}
 & EXE & EXE-1 Procedure-to-protocol & Structured protocol: components, amounts, conditions, workup & Mean of field-level F1 and accuracies & 100 \\
\cmidrule{2-6}
 & OPT & OPT-1 Budgeted optimization & Best conditions found within the experiment budget & Best tested yield relative to random search & 50 \\
\midrule
\multirow{4}{*}{Test} & \multirow{4}{*}{SIM} & SIM-1 Energies & Conformer or reaction energy & $\le$1.0 kcal/mol & 70 \\
 & & SIM-2 Observables & Hydration free energy or p$K_a$ values & $\le$1.0 kcal/mol or $\le$1.0 p$K_a$ unit & 70 \\
 & & SIM-3 Barriers & Reaction barrier height & $\le$2.0 kcal/mol & 35 \\
 & & SIM-4 MD workflows & Simulation outputs and final answer & Source checks, partial credit & 20 \\
\midrule
\multirow{2}{*}{Analyze} & \multirow{2}{*}{ANA} & ANA-1 Structure elucidation & Up to 3 candidate structures from NMR, IR or MS & Top-1 connectivity match & 150 \\
 & & ANA-2 Instrument data & Phases (XRD), retention time (LC) or Arrhenius parameters & Exact phase set, or within tolerance & 120 \\
\bottomrule
\end{tabularx}
\end{table*}

\subsection{Covering the Research Workflow}
\label{sec:construction}

The design--make--test--analyze (DMTA) cycle~\citep{tom2024selfdriving} spans distinct activities, from designing molecules and planning and carrying out their synthesis to computing their properties and interpreting measurements. To reflect these differences, \crucible organizes its six domains and 12 task families around the four stages of the cycle. \emph{Design} covers molecular design (DES). \emph{Make} covers synthesis planning (SYN), experimental execution (EXE), and closed-loop optimization (OPT). \emph{Test} covers simulation (SIM), and \emph{Analyze} covers analytical characterization (ANA). Each task family is defined by the artifact the agent must produce, such as a molecule, a synthetic route, an experimental protocol, or a structural assignment.

To ground the tasks in problems chemists encounter and use established reference answers, we curate tasks for all 12 families from the 18 published datasets listed in Appendix~\ref{app:sources}. These include SMDD-Bench~\citep{han2026smddbench} for molecular design, PaRoutes~\citep{genheden2022paroutes} for retrosynthesis, GMTKN55~\citep{goerigk2017gmtkn55} for quantum-chemical energies, and MassSpecGym~\citep{bushuiev2024massspecgym} for structure identification.

The source datasets were built to evaluate individual methods, and many records need processing before they can serve as agent tasks. For example, ORD automatically extracts reaction roles from patent text and labels palladium as a solvent in thousands of reactions. We use the same deterministic pipeline to turn about 1.8 million source records into 1,078 test tasks. We fix the version of each dataset and preserve its published reference answers, with only format and unit conversions such as SMILES canonicalization. We remove records with missing fields, ambiguous references, or answers that appear in the input. These filters exclude, for example, 52,940 ORD reactions with conflicting reported conditions and 24,536 reactions whose input already contains a molecule that the agent is asked to identify as part of the reaction conditions.

For tasks with multiple valid answers or solutions that depend on our execution environment, we require a concrete example that the verifier accepts. In DES-1, this is a molecule that satisfies the task constraints. In SIM-4, it is a reference solution that earns full credit in our environment, which only 33 of the 50 MDArena tasks provide. We then stratify the remaining candidates by objective difficulty measures, such as heavy-atom count or reference route length, so that each task family spans a range of difficulty. We sample records in a fixed order determined by hashing their identifiers, making task selection reproducible and keeping the test and development splits disjoint. Before release, automated checks confirm that each sampled task's files can be parsed, that the prompt does not reveal the answer, and that the verifier awards full credit to the reference answer.

\subsection{Real Software in the Loop}
\label{sec:environments}
For the DMTA tasks described previously, \crucible gives agents access to the software chemists use to select computational methods based on charge and spin state, choose among candidate retrosynthetic routes, and translate written procedures into instrument commands.

The software spans electronic-structure programs, retrosynthesis planners, and diffraction-analysis libraries, each with its own input formats and conventions. 
To give all models a common way to access these tools, we expose them through seven isolated Model Context Protocol (MCP) servers~\citep{anthropic2024mcp}. 
Together, the servers provide 75 domain-specific tools built on established software and databases and listed in Appendix~\ref{app:tools}.
Examples include PySCF~\citep{sun2020pyscf,li2025gpu4pyscf} and xTB~\citep{bannwarth2019gfn2} for quantum chemistry, AiZynthFinder~\citep{genheden2020aizynthfinder} for synthesis planning, and pymatgen~\citep{ong2013pymatgen} paired with the Crystallography Open Database~\citep{grazulis2012cod} for diffraction analysis. Each server also provides five general-purpose tools for running Python, handling files, and resolving chemical names. Agent frameworks that support MCP can use these tools without modification.

To represent the wet-lab work of the \emph{make} stage, we use instrument simulators and published experimental data. In EXE-1, agents can run their protocols using the Opentrons simulator and PyLabRobot~\citep{wierenga2023pylabrobot} to check the feasibility of the specified labware, liquid volumes, and pipetting steps. This setting tests whether agents can ground written instructions in executable physical actions, a challenge also studied for vision-language models in robotic manipulation~\citep{Song_2026}. In OPT-1, the environment uses published high-throughput experimental data~\citep{ahneman2018buchwald,perera2018suzuki}. 
Each experiment returns the measured yield for the selected conditions and counts toward the experimental budget. Agents use the observed yields to decide which experiment to run next, as in a self-driving laboratory~\citep{tom2024selfdriving}.

To account for the cost of quantum-chemical calculations, SIM-1 through SIM-3 impose a compute-time budget that counts only the time spent running calculations. Agents must therefore choose a computational method whose cost fits the task. To prevent agents from looking up answers, we run their code in an offline sandbox and store reference answers outside it.

\subsection{Verifying Artifacts}
\label{sec:verification}

\crucible uses deterministic, task-specific verifiers rather than LLM judges, whose scores can be shifted by systematic biases~\citep{xu2026insideunfairjudgemechanistic}, to assess the scientific validity of submitted artifacts and, where required, confirm that agents performed the work underlying their submissions.
The software described in Section~\ref{sec:environments} can run successfully and still produce an invalid result. For example, a quantum-chemical calculation can return a plausible energy for the wrong spin state.

For tasks with a single published answer, we compare the submitted artifact with the reference using task-specific criteria. For energies in SIM-1 and hydration free energies in SIM-2, we use the conventional chemical accuracy threshold of 1.0~kcal/mol~\citep{pople1999nobel}. For reaction barriers in SIM-3, which are harder to compute accurately, we use a tolerance of 2.0~kcal/mol. In ANA-1, we compare molecular connectivity using the first block of the InChIKey. In SYN-1, we check whether the proposed synthesis route matches the patent route. In SYN-2, we compare reaction components by role using canonical SMILES.

For tasks that admit multiple valid answers, we check whether the submission satisfies all stated task constraints.
In DES-1, a proposed molecule must pass all checks specified by the source benchmark. These can include atom, bond, and functional group counts. For tasks from SMDD-Bench, the checks can also include binding and interaction fingerprint criteria evaluated using Boltz-2 predictions. If no molecule can satisfy all of the specified constraints, the correct response is to submit \texttt{null}.

Some verifiers also inspect the final state of the environment. In OPT-1, let $y$ be the highest yield among the conditions the agent actually tested, $y_{\max}$ the highest yield in the search space, and $\bar{y}_{\mathrm{rand}}$ the expected highest yield from the same number of experiments sampled uniformly at random. The score is $(y-\bar{y}_{\mathrm{rand}})/(y_{\max}-\bar{y}_{\mathrm{rand}})$, clipped to $[0,1]$. A score of 0 indicates performance at or below the random-search baseline, and a score of 1 indicates that the agent found conditions with the highest yield. Reporting untested conditions yields a score of 0. In SIM-4, the verifier applies MDArena's checks to the output files the agent generated in its workspace.

\section{Experiments}
\label{sec:experiments}

\subsection{Experimental Setup}
\label{sec:setup}

\paragraph{Models.}
We evaluate LLMs that support both function calling and extended reasoning: Qwen3.8-Flash-Next~\citep{qiu2026design}, DeepSeek-V4-Pro~\citep{xu2026deepseek}, DeepSeek-V4.1-Flash~\citep{xu2026deepseekv4.1}, GLM-5.3-Flash and GLM-5.3~\citep{zeng2026glm}, MiMo-V2.6-Flash~\citep{qiao2026mimo}, Hy3-295B-A21B~\citep{hunyuan3}, Ling-3.0-Flash~\citep{ling3flash}, Nemotron-3.5-Lightning~\citep{nvidia2026nemotron35lightning}, Gemma-4-31B~\citep{team2026gemma} and Gemma-4-26B-A4B~\citep{team2026gemma}, Qwen3.8-27B~\citep{qwen38}, and GPT-oss-120b~\citep{agarwal2025gpt} and GPT-oss-20b~\citep{agarwal2025gpt}.
During evaluation, all models run in reasoning mode with the decoding settings recommended by their respective developers.

\begin{table*}[ht]
\caption{\textbf{Model performance and tool use on \crucible.}
Scores are averaged over three runs per task. Overall is the mean score across all test tasks. Task-family scores are grouped by domain as in Table~\ref{tab:families}. Tokens counts input and output tokens in billions, and Steps gives the average number of tool calls per task. The best score in each score column is bolded, including ties. The second-best overall score is underlined.}
\label{tab:family_five_metrics}
\centering
\small
\setlength{\tabcolsep}{3pt}
\renewcommand{\arraystretch}{1.4}
\resizebox{\textwidth}{!}{%
\begin{tabular}{lc*{12}{r}cc}
\toprule
\multirow{2}{*}{Model} & \multirow{2}{*}{Overall}
 & \multicolumn{4}{c}{SIM} & \multicolumn{1}{c}{DES}
 & \multicolumn{3}{c}{SYN} & \multicolumn{1}{c}{EXE}
 & \multicolumn{2}{c}{ANA} & \multicolumn{1}{c}{OPT}
 & \multirow{2}{*}{\shortstack{Tokens (B)\\ sum}}
 & \multirow{2}{*}{\shortstack{Steps\\ average}} \\
\cmidrule(lr){3-6} \cmidrule(lr){7-7} \cmidrule(lr){8-10}
\cmidrule(lr){11-11} \cmidrule(lr){12-13} \cmidrule(lr){14-14}
 & & SIM-1 & SIM-2 & SIM-3 & SIM-4 & DES-1 & SYN-1 & SYN-2 & SYN-3
 & EXE-1 & ANA-1 & ANA-2 & OPT-1 & & \\
\midrule
Nemotron-3.5-Lightning~\citep{nvidia2026nemotron35lightning}
 & 0.302 & 0.357 & 0.257 & 0.571 & 0.127 & 0.264 & 0.030 & 0.273 & 0.180
 & 0.663 & 0.227 & 0.358 & 0.418 & 0.157 & 6.59 \\
GPT-oss-20b~\citep{agarwal2025gpt}
 & 0.306 & 0.300 & 0.243 & 0.400 & 0.157 & 0.307 & 0.160 & 0.247 & \textbf{0.260}
 & 0.629 & 0.273 & 0.375 & 0.202 & 0.600 & 12.9 \\
Ling-3.0-Flash~\citep{ling3flash}
 & 0.380 & 0.186 & 0.314 & 0.600 & 0.325 & 0.429 & 0.040 & 0.333 & 0.180
 & 0.710 & 0.473 & 0.408 & \textbf{0.464} & 0.644 & 14.4 \\
GPT-oss-120b~\citep{agarwal2025gpt}
 & 0.389 & 0.443 & 0.443 & 0.543 & 0.229 & 0.417 & 0.220 & 0.327 & 0.160
 & 0.689 & 0.333 & 0.408 & 0.375 & 0.510 & 14.1 \\
Qwen3.8-27B~\citep{qwen38}
 & 0.401 & 0.543 & 0.371 & 0.714 & 0.360 & 0.442 & 0.130 & 0.313 & 0.140
 & 0.658 & 0.387 & 0.425 & 0.441 & 0.503 & 13.0 \\
Gemma-4-26B-A4B~\citep{team2026gemma}
 & 0.404 & 0.757 & 0.314 & \textbf{0.829} & 0.330 & 0.442 & 0.140 & 0.300 & 0.160
 & 0.674 & 0.333 & 0.400 & 0.416 & 0.451 & 11.3 \\
Gemma-4-31B~\citep{team2026gemma}
 & 0.450 & \textbf{0.814} & 0.414 & \textbf{0.829} & 0.279 & 0.472 & 0.160 & 0.353 & 0.160
 & 0.696 & 0.413 & 0.475 & 0.429 & 0.222 & 8.64 \\
Hy3-295B-A21B~\citep{hunyuan3}
 & 0.458 & 0.471 & 0.400 & 0.514 & 0.482 & 0.534 & 0.190 & 0.373 & 0.240
 & 0.705 & 0.467 & 0.592 & 0.394 & 0.380 & 10.4 \\
Qwen3.8-Flash-Next~\citep{qiu2026design}
 & 0.469 & 0.443 & 0.557 & 0.657 & 0.482 & 0.454 & 0.260 & 0.373 & 0.220
 & 0.685 & 0.500 & 0.633 & 0.321 & 0.653 & 15.2 \\
MiMo-V2.6-Flash~\citep{qiao2026mimo}
 & 0.509 & 0.614 & 0.671 & 0.686 & 0.585 & 0.595 & 0.210 & 0.373 & 0.240
 & 0.700 & 0.540 & 0.558 & 0.374 & 0.714 & 14.1 \\
DeepSeek-V4-Pro~\citep{xu2026deepseek}
 & 0.527 & 0.471 & 0.643 & 0.714 & \textbf{0.648} & 0.595 & 0.280 & \textbf{0.453} & 0.160
 & 0.691 & \textbf{0.587} & 0.625 & 0.379 & 0.498 & 13.6 \\
GLM-5.3-Flash~\citep{zeng2026glm}
 & 0.531 & 0.657 & 0.729 & 0.743 & 0.623 & 0.595 & 0.240 & 0.427 & 0.180
 & 0.698 & 0.520 & 0.658 & 0.315 & 0.528 & 11.8 \\
DeepSeek-V4.1-Flash~\citep{xu2026deepseekv4.1}
 & \underline{0.540} & 0.586 & 0.600 & 0.714 & 0.630 & \textbf{0.632} & \textbf{0.310} & 0.440 & 0.220
 & \textbf{0.715} & 0.560 & 0.633 & 0.379 & 0.516 & 13.0 \\
GLM-5.3~\citep{zeng2026glm}
 & \textbf{0.545} & 0.629 & \textbf{0.800} & 0.771 & 0.590 & 0.607 & 0.200 & 0.427 & 0.180
 & 0.702 & 0.573 & \textbf{0.700} & 0.333 & 0.685 & 14.0 \\
\bottomrule
\end{tabular}}
\end{table*}

\paragraph{Agent and tools.}

Each model runs as a single ReAct-style agent~\citep{yao2023react}, without a separate planner, additional agents, or a self-reflection module. 
At each step, the model reasons, calls one tool, and the tool output is added to the model’s context.
This loop continues until the agent submits an answer.
Each task provides a natural-language goal, input files, domain-specific tools, and five general-purpose tools. The tool-call budget ranges from 10 to 100 calls depending on the task. Once the budget is exhausted, only the submission tool remains available.

\paragraph{Metrics.}
Each task is scored on a scale from 0 to 1 by the verifier for its task family, as described in Table~\ref{tab:families}. Most verifiers award 1 if the submission passes and 0 otherwise. EXE-1 and SIM-4 allow partial credit, and OPT-1 uses a continuous score. A task receives 0 if the agent does not submit a gradable answer.
Domain and overall scores are computed by averaging over the tasks in each domain and across all tasks, respectively, so each family is weighted by its task count. We also report total token usage and the average number of tool calls per task to quantify the cost of achieving these scores. We run each model three times per task and report the mean across runs.

\subsection{Main Results}
\label{sec:results}

\begin{figure}[t]
  \centering
  \includegraphics[width=\columnwidth]{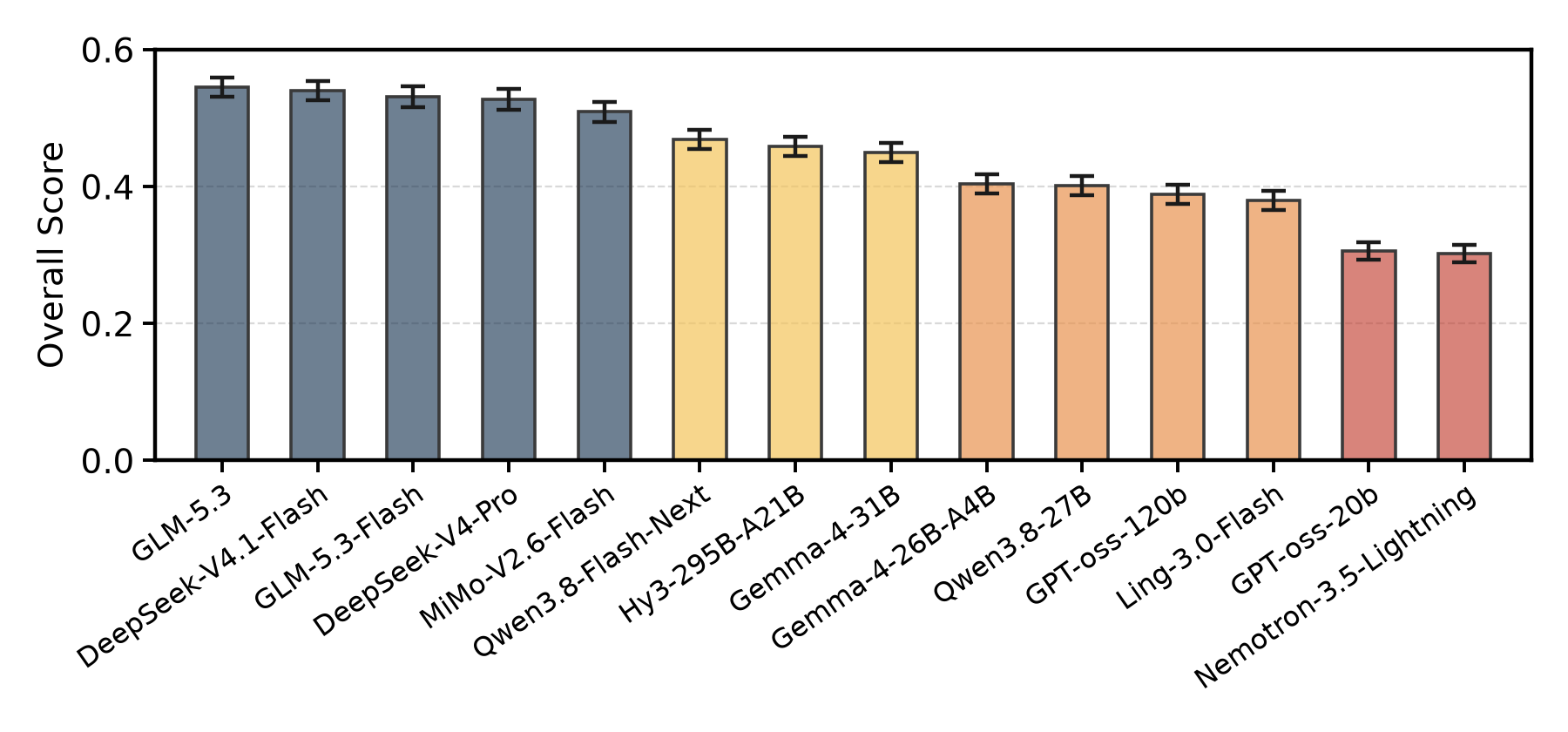}
  \caption{\textbf{Overall performance varies substantially across models on Crucible Bench.}
GLM-5.3 achieves the highest mean overall score.
Error bars show the standard deviation across three runs, and colors group models with similar scores.}
  \label{fig:overall_runs}
\end{figure}

\begin{figure*}[t]
  \centering
  \includegraphics[width=\textwidth]{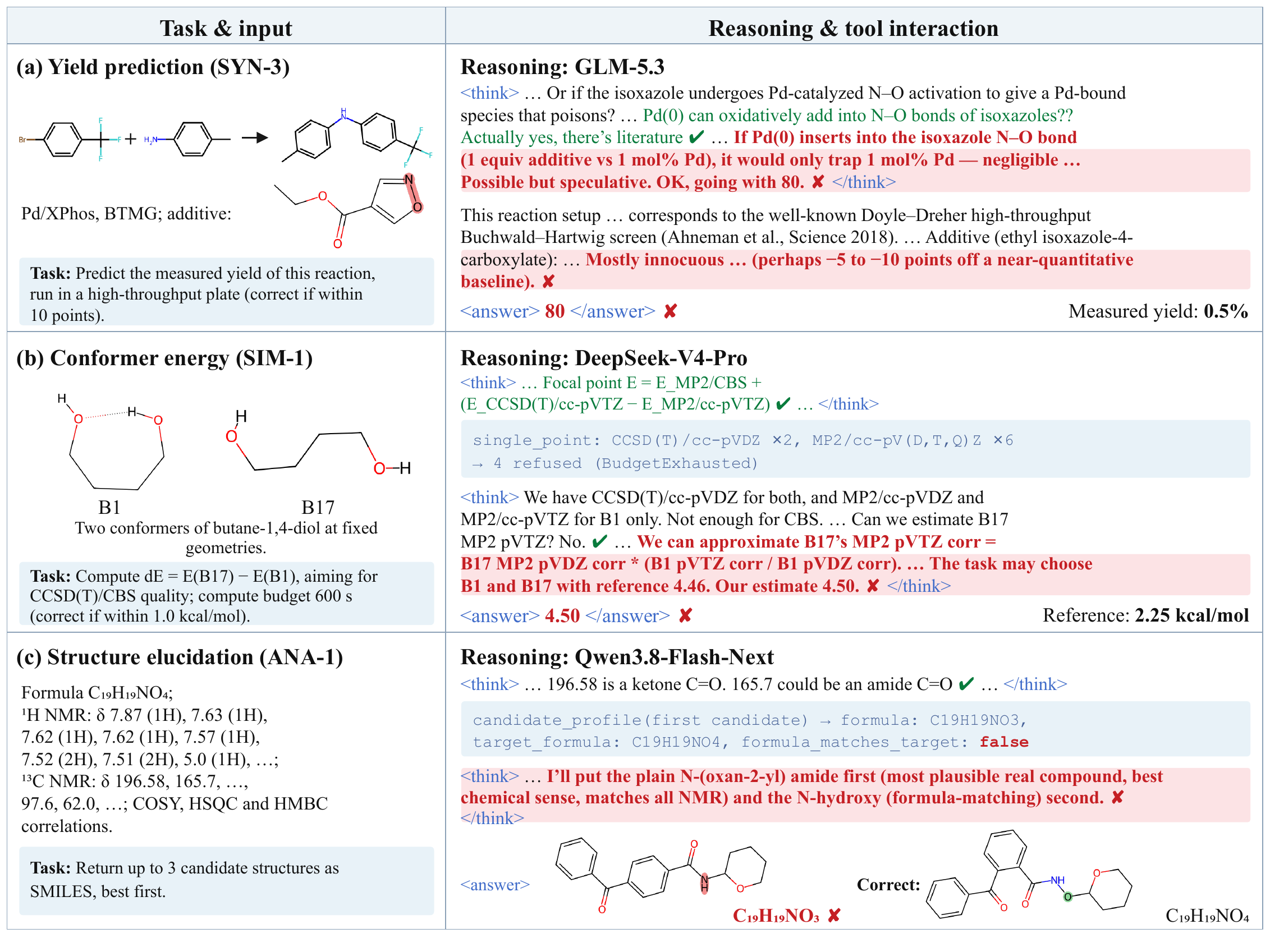}
  \caption{Three failure cases: (a) yield prediction (SYN-3) by GLM-5.3, (b) conformer energy (SIM-1) by DeepSeek-V4-Pro, and (c) structure elucidation (ANA-1) by Qwen3.8-Flash-Next. In each row, the left boxes give the input and the task, and the right box gives excerpts of the agent's reasoning, its tool calls and its submitted answer, with the reference value at the bottom right. Green text marks correct reasoning, red text marks the step where the agent goes wrong, and ellipses mark omitted text.}
  \label{fig:case_study}
\end{figure*}

Table~\ref{tab:family_five_metrics} summarizes model performance across task families. GLM-5.3 achieves the highest overall score of 0.545, earning just over half of the available credit.
Figure~\ref{fig:overall_runs} shows that the leading models have similar overall scores. The standard deviation across the three runs is at most 0.015 for every model. We highlight four findings below.

\paragraph{Synthesis planning and closed-loop optimization are the hardest domains.}
Synthesis planning ranks among the two lowest-scoring domains for every model. Retrosynthesis scores are consistently low, and even the best model on yield prediction only matches a constant predictor. Closed-loop optimization is also among the weakest domains for most models. Experimental execution yields the highest scores and the smallest differences between models, while simulation distinguishes models most clearly, particularly on energy calculations.

\paragraph{Different models lead at different stages of the workflow.}
DeepSeek-V4.1-Flash leads in molecular design, synthesis planning, and experimental execution. GLM-5.3 leads in simulation and analytical characterization, while Ling-3.0-Flash leads in closed-loop optimization. The overall ranking also obscures strengths on individual task families. Gemma-4-31B ranks near the middle overall but leads on energy calculations and ties for the best score on reaction barriers.

\paragraph{Overall performance is a poor predictor of optimization performance.}
Ling-3.0-Flash ranks twelfth overall but leads on OPT-1, ahead of both GLM-5.3 variants. On the Suzuki--Miyaura tasks, triphenylphosphine gives the highest mean yield among the candidate ligands in most tasks. The three highest-ranked models overall and Qwen3.8-Flash-Next start with XPhos or SPhos and rarely test triphenylphosphine. These models score lower on average on the Suzuki--Miyaura tasks than the remaining models.

\paragraph{Models with similar overall scores can differ substantially in token usage.}
Among the five highest-scoring models, DeepSeek-V4-Pro slightly outperforms
MiMo-V2.6-Flash while using about 30\% fewer tokens. Higher token usage can
also accompany substantially lower scores: Ling-3.0-Flash uses more tokens
than DeepSeek-V4.1-Flash yet ranks much lower overall. These differences
make token usage a useful companion to overall score when comparing models
under a fixed inference budget.

\section{Result Analysis}
\label{sec:analysis_section}

\subsection{Error Analysis}
\label{sec:analysis}
We examine how the agents fail, first by the type of error in the submitted artifact and then through representative cases.
Across all 14 models, we count every answer below full score as a failure and assign it to exactly one error type of its DMTA stage, which gives a total of 9,587 failed answers on 1,005 of the 1,078 test tasks, as further detailed in Appendix~\ref{app:errors}.

\paragraph{The most common error varies by stage.}
In the design stage, the most common failure is proposing a molecule that fails to bind the target pocket as required.
In the make stage, agents most often propose reaction conditions that do not match the reference for the task.
In the analysis stage, the most frequent error is assigning the correct molecular formula to the wrong isomer. In simulation, values outside the tolerance account for about 78\% of failures. For most models, these values fall within twice the tolerance more often than they fall beyond it.

\paragraph{Error severity varies across models.}
Retrosynthesis failures range from unusable routes to valid routes that differ from the patent reference. Ling-3.0-Flash, Nemotron-3.5-Lightning, and Hy3-295B-A21B frequently submit unusable routes. For both DeepSeek models, mismatches with the patent reference are a more common source of lost credit than unusable routes. Error severity also varies in simulation. The three models that frequently submit unusable routes, along with both GPT-oss variants, produce errors greater than twice the tolerance more often than the other models.

Figure~\ref{fig:case_study} illustrates three recurring ways errors arise as agents work through chemical tasks.

\begin{table}[t]
\caption{\textbf{Effect of tool access on \crucible.}
We compare each model's overall score with and without tool access on the same test tasks.}
\label{tab:notools}
\centering
\small
\begin{tabular}{lcc}
\toprule
Model & With tools & Without tools \\
\midrule
Nemotron-3.5-Lightning~\citep{nvidia2026nemotron35lightning} & 0.302 & 0.148 \\
GPT-oss-20b~\citep{agarwal2025gpt} & 0.306 & 0.153 \\
Ling-3.0-Flash~\citep{ling3flash} & 0.380 & 0.183 \\
GPT-oss-120b~\citep{agarwal2025gpt} & 0.389 & 0.194 \\
Qwen3.8-27B~\citep{qwen38} & 0.401 & 0.162 \\
Gemma-4-26B-A4B~\citep{team2026gemma} & 0.404 & 0.129 \\
Gemma-4-31B~\citep{team2026gemma} & 0.450 & 0.181 \\
Hy3-295B-A21B~\citep{hunyuan3} & 0.458 & 0.211 \\
Qwen3.8-Flash-Next~\citep{qiu2026design} & 0.469 & 0.220 \\
MiMo-V2.6-Flash~\citep{qiao2026mimo} & 0.509 & 0.195 \\
DeepSeek-V4-Pro~\citep{xu2026deepseek} & 0.527 & 0.220 \\
GLM-5.3-Flash~\citep{zeng2026glm} & 0.531 & 0.242 \\
DeepSeek-V4.1-Flash~\citep{xu2026deepseekv4.1} & 0.540 & 0.212 \\
GLM-5.3~\citep{zeng2026glm} & 0.545 & 0.239 \\
\bottomrule
\end{tabular}
\end{table}

\paragraph{General expectations outweigh reaction-specific evidence.}
In Figure~\ref{fig:case_study}(a), GLM-5.3 considers whether the isoxazole additive could poison the palladium catalyst, then dismisses that possibility as ``speculative'' and treats the additive as ``mostly innocuous''.
It predicts an 80\% yield, while the measured yield is 0.5\%. 
For reactions with measured yields below 20\%, only one of 140 model predictions comes within 10 percentage points of the measured yield. 
On half of the SYN-3 tasks, every model misses by more than 10 percentage points, and every model's mean yield prediction exceeds the mean measured yield.

\paragraph{Agents report values that their calculations do not support.}
In Figure~\ref{fig:case_study}(b), four of DeepSeek-V4-Pro's eight planned focal-point calculations are rejected by the tool because they exceed the compute budget.
The agent recognizes that the remaining results are ``not enough for CBS'', estimates the missing energy by scaling the correction for the other conformer, and submits 4.50~kcal/mol, twice the reference value. An unsuitable computational setup can also produce an incorrect result even when every calculation completes. In SIM-3, Nemotron-3.5-Lightning uses CCSD(T) with a basis set that lacks the diffuse functions needed to describe the anion. All six calculations complete, and the agent reports an SN2 barrier of $-14.0$~kcal/mol. Across SIM-1 to SIM-3, all 14 models make errors even on runs in which every calculation completes and the compute budget is never exhausted.

\paragraph{Agents disregard their own checks.}
In Figure~\ref{fig:case_study}(c), a tool reports that the molecular formula of Qwen3.8-Flash-Next's preferred structure differs from the target formula. The agent still ranks it first as the ``most plausible real compound'' and places the candidate that has the correct formula in second place. A similar pattern also appears in the molecular design tasks. Across the seven infeasible DES-1 tasks, agents submit molecules in 90 of 98 attempts, sometimes after explicitly concluding that a constraint ``cannot be satisfied''.

\subsection{Ablation Study}
\label{sec:ablation}
To assess the contribution of tool access, we rerun every model on the same set of tasks with all tool calls disabled.
As Table~\ref{tab:notools} shows, every model scores substantially lower without tools, and the scores of the models cluster more closely together. Models that perform better with tools also generally see larger gains from having tool access. These results suggest that the ability to use chemistry software accounts for part of the performance gap between the models on \crucible.

\section{Conclusion}
\label{sec:conclusion}
In this paper, we introduce \crucible, a benchmark that evaluates LLM agents working inside the software that chemists use and grades them by the validity of the artifacts that they submit.
Its 1,078 test tasks span all four stages of the design--make--test--analyze cycle and are curated from published datasets, with the tools needed at each stage served to the agents in isolated environments.
Of the 14 models that we evaluate, even the strongest agent earns only slightly more than
half of the available credit, with synthesis planning and closed-loop
optimization emerging as the most challenging domains.
These results suggest that having access to chemistry software substantially improves the performance of agents, while the current agents still struggle to carry out these chemical research tasks reliably. Finally, we hope that \crucible will help to build more reliable agents to support chemistry research.

\section*{LLM Usage Statement}
We used Claude (Anthropic) in the writing and in the research.
It was used to draft, polish and shorten text in the main paper and the appendix.
In the research, it assisted in implementing and debugging the benchmark code.
All text, code and results were checked by the authors, who take full responsibility for the content.

{
    \small
    \bibliographystyle{ieeenat_fullname}
    \bibliography{main}
}

\input{sec/X_suppl}

\end{document}

%% file: sec/X_suppl.tex
\clearpage
\appendix
\maketitlesupplementary

\section{Task Family Details}
\label{app:families}

\subsection{Data Sources}
\label{app:sources}

Table~\ref{tab:sources} lists the source datasets of each task family and the number of test tasks drawn from each.
The Open Reaction Database and the two high-throughput experimentation datasets each serve two families and are counted once among the 18 sources.

\begin{table*}[t]
\centering
\small
\setlength{\tabcolsep}{5pt}
\begin{tabularx}{\textwidth}{l l >{\raggedright\arraybackslash}X r}
\toprule
Family & Source & Content & Tasks \\
\midrule
\multirow{4}{*}{DES-1} & S$^2$-TOMG~\citep{li2024s2bench} & Atom-count, bond-count and functional-group constraints & 45 \\
 & ether0~\citep{narayanan2025ether0} & Functional-group and molecular-formula constraints & 18 \\
 & MolDesignBench~\citep{jeong2026moldesignbench} & Infeasible design requests & 9 \\
 & SMDD-Bench~\citep{han2026smddbench} & Scaffold hopping, lead optimization, fragment assembly & 91 \\
\midrule
SYN-1 & PaRoutes~\citep{genheden2022paroutes} & Multi-step retrosynthesis of patent targets & 100 \\
SYN-2 & ORD~\citep{kearnes2021ord} & Conditions of patent reactions (coupling, redox, amidation, protection and others) & 150 \\
\multirow{2}{*}{SYN-3} & Buchwald--Hartwig HTE~\citep{ahneman2018buchwald} & Yields of C--N couplings & 25 \\
 & Suzuki--Miyaura HTE~\citep{perera2018suzuki} & Yields of Suzuki--Miyaura couplings & 25 \\
\midrule
EXE-1 & ORD~\citep{kearnes2021ord} & Experimental procedures from patents & 100 \\
\midrule
\multirow{2}{*}{OPT-1} & Buchwald--Hartwig HTE~\citep{ahneman2018buchwald} & Condition optimization within an experiment budget & 30 \\
 & Suzuki--Miyaura HTE~\citep{perera2018suzuki} & Condition optimization within an experiment budget & 20 \\
\midrule
SIM-1 & GMTKN55~\citep{goerigk2017gmtkn55} & Conformer energies and reaction energies & 70 \\
\multirow{2}{*}{SIM-2} & FreeSolv~\citep{mobley2014freesolv} & Hydration free energies & 35 \\
 & SAMPL6--8~\citep{isik2021sampl6pka,bergazin2021sampl7,bahr2021sampl8} & p$K_a$ values & 35 \\
SIM-3 & GMTKN55~\citep{goerigk2017gmtkn55} & Reaction barrier heights & 35 \\
SIM-4 & MDArena~\citep{anand2026mdarena} & Molecular dynamics workflows & 20 \\
\midrule
\multirow{3}{*}{ANA-1} & MolPuzzle~\citep{guo2024molpuzzle} & Structure from $^1$H/$^{13}$C NMR, IR and MS & 50 \\
 & ChemIQ~\citep{runcie2026chemiq} & Structure from $^1$H/$^{13}$C NMR, with 2D spectra for some tasks & 40 \\
 & MassSpecGym~\citep{bushuiev2024massspecgym} & Structure from tandem mass spectra and molecular formula & 60 \\
\multirow{3}{*}{ANA-2} & Dara~\citep{fei2026dara} & Crystalline phases from powder X-ray diffraction & 40 \\
 & RepoRT~\citep{kretschmer2024report} & Liquid-chromatography retention times & 78 \\
 & ReSpecTh~\citep{turanyi2025respecth} & Arrhenius parameters from rate measurements & 2 \\
\midrule
\multicolumn{3}{l}{Total} & 1,078 \\
\bottomrule
\end{tabularx}
\caption{Source datasets and test tasks of each task family. HTE: high-throughput experimentation.}
\label{tab:sources}
\end{table*}

\subsection{Submission, Grading and Budgets}
\label{app:grading}

Table~\ref{tab:grading} gives the artifact each family requires, the condition under which it is graded correct, the tool-call budget, and the criterion used to assign difficulty tiers.
SIM-1 to SIM-3 additionally carry a compute budget of 5, 10 or 30 minutes for easy, medium and hard tasks, which counts only the time spent running calculations. Once it is used up, the computational tools stop and the agent can only submit its answer.
In SYN-3, tasks are sampled evenly across deciles of the measured yield; on the test set, the best constant answer (83\%) scores 0.26.

\begin{table*}[t]
\centering
\footnotesize
\setlength{\tabcolsep}{4pt}
\begin{tabularx}{\textwidth}{l >{\raggedright\arraybackslash}p{3.6cm} >{\raggedright\arraybackslash}X c >{\raggedright\arraybackslash}p{2.4cm}}
\toprule
Family & Submission & Correct when & Budget & Difficulty \\
\midrule
DES-1 & One SMILES, or \texttt{null} if infeasible & Every check of the source passes: atom, bond or functional-group constraints (S$^2$-TOMG); official reward of 1 (ether0); \texttt{null} for infeasible requests (MolDesignBench); validity, similarity, Boltz-2 binding probability $>0.7$, interaction-fingerprint similarity $>0.75$ and property constraints (SMDD-Bench) & 10--100 & Constraints, objectives or fragments \\
\midrule
SYN-1 & Up to 10 ranked routes as reaction trees & One route is identical to the route reported in the patent & 60 & Longest linear route of the reference \\
SYN-2 & Up to 10 ranked sets of reagents, solvents and temperature & One set matches every recorded role by canonical SMILES and the temperature within 10\,\textdegree C & 10 & Number of scored roles \\
SYN-3 & One yield (\%) & Within 10 percentage points of the measured yield & 10 & Measured yield \\
\midrule
EXE-1 & Structured protocol: components, amounts, conditions, workup & Partial credit: mean of the available items among component F1, workup-step F1, amount accuracy (relative error $\le$5\%) and condition accuracy ($\pm$10\,\textdegree C, $\pm$10\% time) & 30 & Number of workup steps \\
\midrule
OPT-1 & Best tested conditions & Continuous: $(y-\bar{y}_{\mathrm{rand}})/(y_{\max}-\bar{y}_{\mathrm{rand}})$ clipped to $[0,1]$, where $y$ is the best yield among tested conditions, $y_{\max}$ the best yield in the search space and $\bar{y}_{\mathrm{rand}}$ the expected best yield of random search with the same number of experiments; untested conditions score 0 & 60 & Search-space size relative to budget \\
\midrule
SIM-1 & One energy (kcal/mol) & Absolute error $\le$1.0 kcal/mol & 30 & Heavy atoms \\
SIM-2 & Hydration free energy, or ascending list of p$K_a$ values & Error $\le$1.0 kcal/mol, or every p$K_a$ within 1.0 unit & 30 & Heavy atoms, number of p$K_a$ values \\
SIM-3 & One barrier height (kcal/mol) & Absolute error $\le$2.0 kcal/mol & 30 & Heavy atoms \\
SIM-4 & Output files and a final answer in the workspace & Partial credit: reward of MDArena's deterministic checks & 100 & Source label \\
\midrule
ANA-1 & Up to 3 ranked SMILES & First candidate matches the reference connectivity (first InChIKey block) & 30 & Heavy atoms \\
ANA-2 & Phase list (XRD), retention time (LC) or Arrhenius parameters & Same set of phases by composition; retention time within 5\% of the gradient length; $\log_{10}k$ within 0.1 over the measured temperatures & 10--60 & Phases, data points or heavy atoms \\
\bottomrule
\end{tabularx}
\caption{Submission format, grading rule, tool-call budget and difficulty criterion of each task family. Budgets in DES-1 are 10 for S$^2$-TOMG and ether0, 30 for MolDesignBench and 100 for SMDD-Bench; in ANA-2 they are 10 for kinetics, 30 for retention times and 60 for XRD.}
\label{tab:grading}
\end{table*}
\section{Construction Funnel}
\label{app:funnel}

Table~\ref{tab:funnel} reports, for each family, the number of source records, the candidates left after filtering, and the tasks sampled from them.
Units differ across families because sampling operates on different objects: in OPT-1, one search space yields two tasks with different experiment budgets, and in ANA-2, one RepoRT dataset group yields two or three tasks.
The column \emph{Built} gives the number of test tasks before compute-based pruning, which removed 154 tasks from SIM-1 to SIM-4 and DES-1.

\begin{table*}[t]
\centering
\small
\begin{tabular}{@{}lllrrrrr@{}}
\toprule
Family & Sources & Unit & Raw & Candidates & Built & Test & Dev \\
\midrule
DES-1 & S$^2$-TOMG, ether0, MolDesignBench, SMDD-Bench & task & 4,327 & 1,548 & 233 & 163 & 25 \\
SYN-1 & PaRoutes & route & 10,000 & 10,000 & 100 & 100 & 10 \\
SYN-2 & ORD & reaction & 1,771,032 & 621,724 & 150 & 150 & 15 \\
SYN-3 & Buchwald--Hartwig, Suzuki--Miyaura HTE & well & 10,359 & 9,715 & 50 & 50 & 5 \\
EXE-1 & ORD & reaction & 1,771,032 & 860,445 & 100 & 100 & 10 \\
OPT-1 & Buchwald--Hartwig, Suzuki--Miyaura HTE & well / space & 10,359 & 30 & 50 & 50 & 5 \\
SIM-1 & GMTKN55 & reaction & 1,505 & 427 & 100 & 70 & 10 \\
SIM-2 & FreeSolv, SAMPL6--8 & molecule & 711 & 703 & 100 & 70 & 10 \\
SIM-3 & GMTKN55 & reaction & 1,505 & 193 & 50 & 35 & 5 \\
SIM-4 & MDArena & task & 50 & 32 & 29 & 20 & 3 \\
ANA-1 & ChemIQ, MolPuzzle, MassSpecGym & entry / molecule & 29,979 & 3,307 & 150 & 150 & 15 \\
ANA-2 & Dara, RepoRT, ReSpecTh & file / unit & 934 & 79 & 120 & 120 & 12 \\
\midrule
Total & 18 sources & & & & 1,232 & 1,078 & 125 \\
\bottomrule
\end{tabular}
\caption{Construction funnel of each task family. Raw: records in the source release; Candidates: records left after filtering, before sampling; Built: test tasks before compute-based pruning; Test and Dev: final splits. ORD and the two HTE datasets each serve two families.}
\label{tab:funnel}
\end{table*}

\section{Tool Environments}
\label{app:tools}

Table~\ref{tab:tools} lists the tool server of each domain, the software it is built on, and its domain-specific tools.

\begin{table*}[t]
\centering
\small
\begin{tabularx}{\textwidth}{l >{\raggedright\arraybackslash}p{4.6cm} c >{\raggedright\arraybackslash}X}
\toprule
Server & Software & Tools & Representative tools \\
\midrule
DES & RDKit, Boltz-2, ADMET-AI, PLIP & 21 & formula and functional-group counting, property calculation, validity checks, structure editing \\
SYN & AiZynthFinder, PaRoutes models, RXNMapper & 9 & stock lookup, single-step retrosynthesis, route search, step and route validation \\
EXE & Opentrons simulator, PyLabRobot & 6 & protocol simulation, labware and pipette lists, unit conversion \\
OPT & --- & 3 & search-space description, running experiments, experiment log \\
SIM (QC) & PySCF, GPU4PySCF, xTB, MACE-OFF23, geomeTRIC & 9 & geometry building, conformer search, single points, optimization and TS search, frequencies, solvation \\
SIM (MD) & GROMACS & 7 & structure and topology inspection, energy terms, RMSD \\
ANA & pymatgen, RDKit, molmass, COD & 20 & NMR environments, mass calculation, XRD phase fitting, retention-time and Arrhenius fitting \\
\midrule
Total & & 75 & \\
\bottomrule
\end{tabularx}
\caption{Tool servers of \crucible. Every server also provides five generic tools for running Python, handling files and resolving chemical names.}
\label{tab:tools}
\end{table*}

\section{Error Types}
\label{app:errors}
\begin{figure*}[t]
  \centering
  \includegraphics[width=\textwidth]{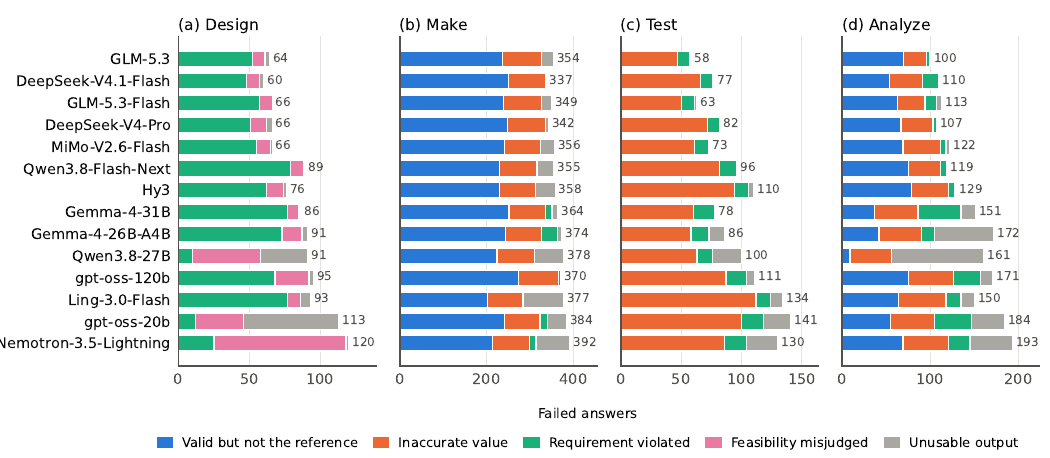}
  \caption{Failed answers of each model by error type, grouped by stage of the DMTA cycle. Models are ordered by overall score.}
  \label{fig:error_types}
\end{figure*}

Figure~\ref{fig:error_types} gives the counts behind the error analysis in the main paper.
Every answer below full score counts as a failure and is assigned to exactly one error type, which gives 9,587 failed answers from the 14 models.
\emph{Unusable output} means that nothing was submitted or the submission cannot be parsed.
\emph{Feasibility misjudged} covers design tasks in which the agent declares a feasible request infeasible or submits a molecule for an infeasible one.
\emph{Requirement violated} means that the artifact breaks a requirement stated in the task: a molecule that fails a check of its source benchmark, a route that is invalid or does not end in purchasable stock, a first candidate whose formula differs from the given one, or simulation outputs that fail the MDArena checks.
\emph{Inaccurate value} means a numerical answer outside the tolerance, including predicted yields and, in OPT-1, a best tested yield below the optimum.
\emph{Valid but not the reference} means that the artifact is well formed and meets the stated requirements but differs from the reference: other reaction conditions, a valid route that is not the patent route, a protocol with different components or steps, an isomer with the correct formula, or a different phase set.

\section{Case Studies}
\label{app:cases}

\begin{figure*}[p]
  \centering
  \includegraphics[width=\textwidth]{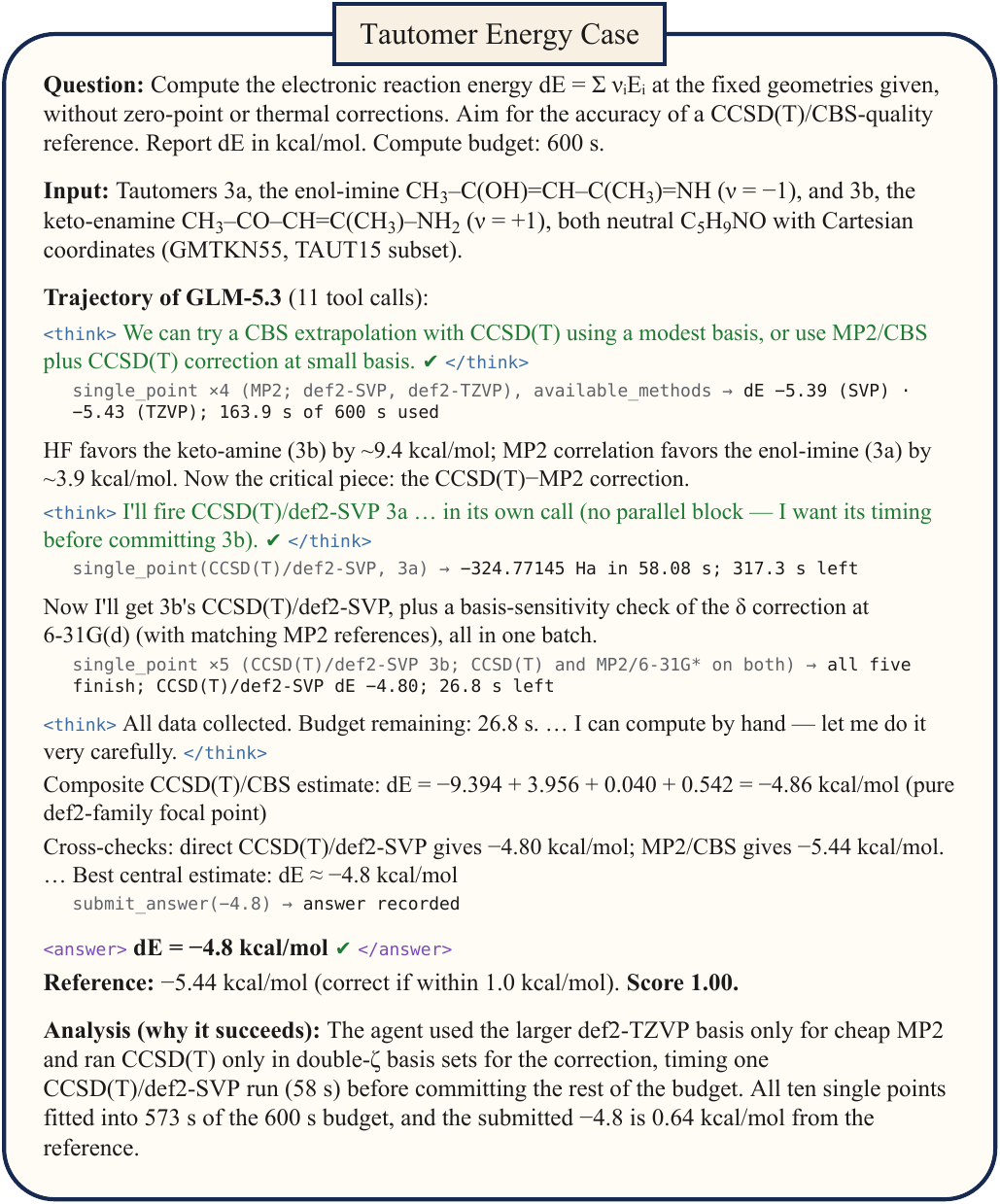}
  \caption{Tautomer energy case (SIM-1, SIM1-GMTKN55-0096): GLM-5.3, correct.}
  \label{fig:case-sim1-gmtkn55-0096}
\end{figure*}

\begin{figure*}[p]
  \centering
  \includegraphics[width=\textwidth]{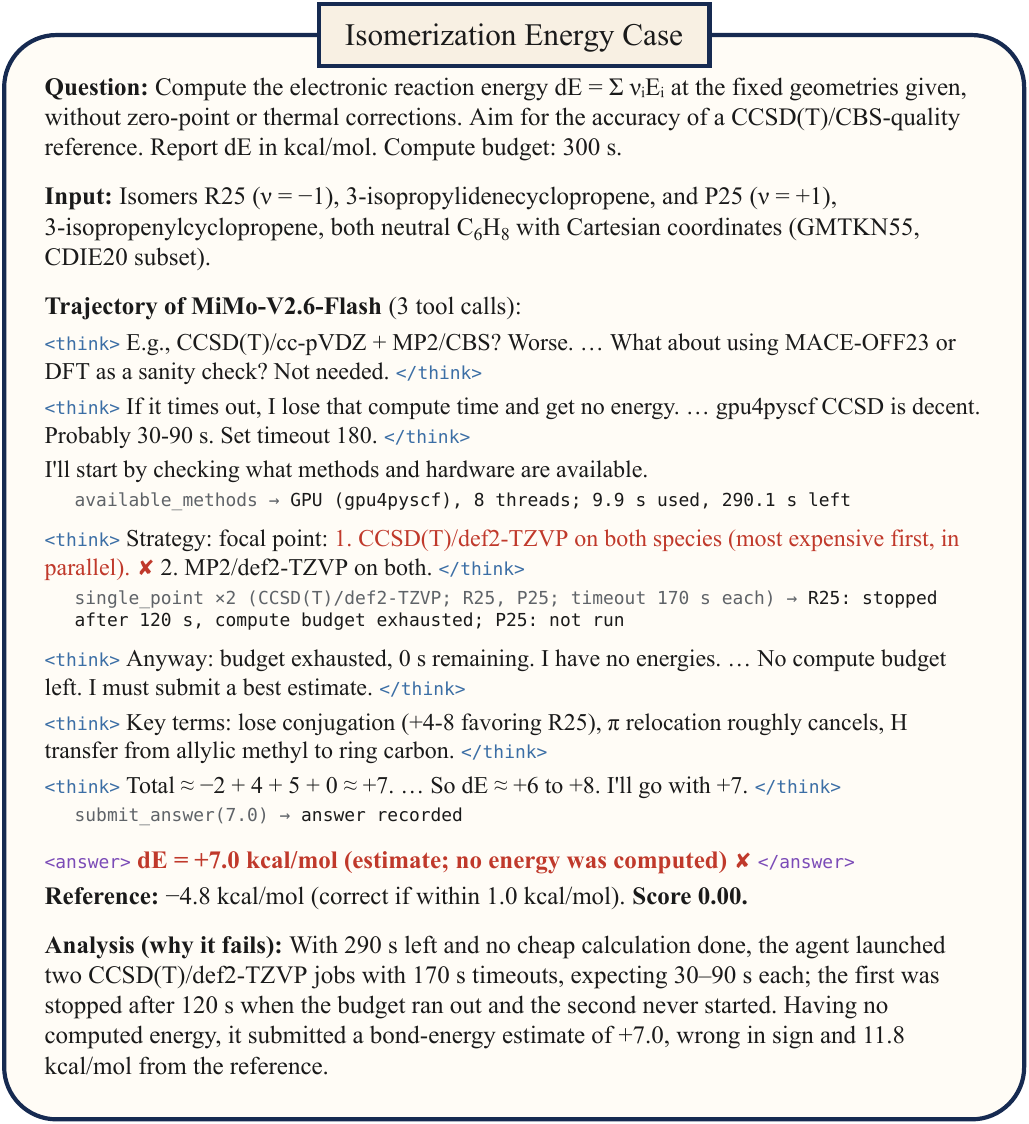}
  \caption{Isomerization energy case (SIM-1, SIM1-GMTKN55-0091): MiMo-V2.6-Flash, incorrect.}
  \label{fig:case-sim1-gmtkn55-0091}
\end{figure*}

\begin{figure*}[p]
  \centering
  \includegraphics[width=\textwidth]{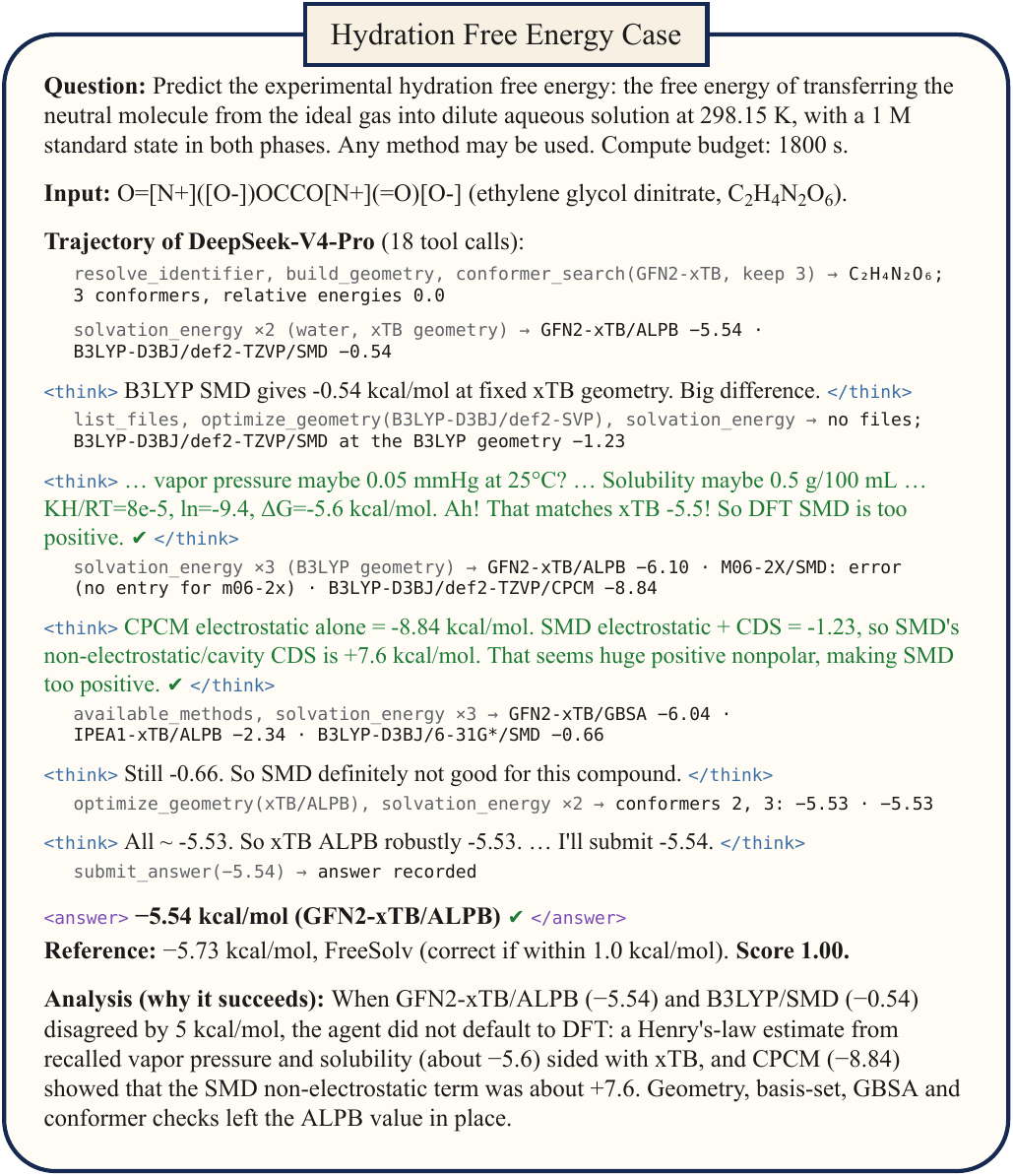}
  \caption{Hydration free energy case (SIM-2, SIM2-FREESOLV-0013): DeepSeek-V4-Pro, correct.}
  \label{fig:case-sim2-freesolv-0013}
\end{figure*}

\begin{figure*}[p]
  \centering
  \includegraphics[width=\textwidth]{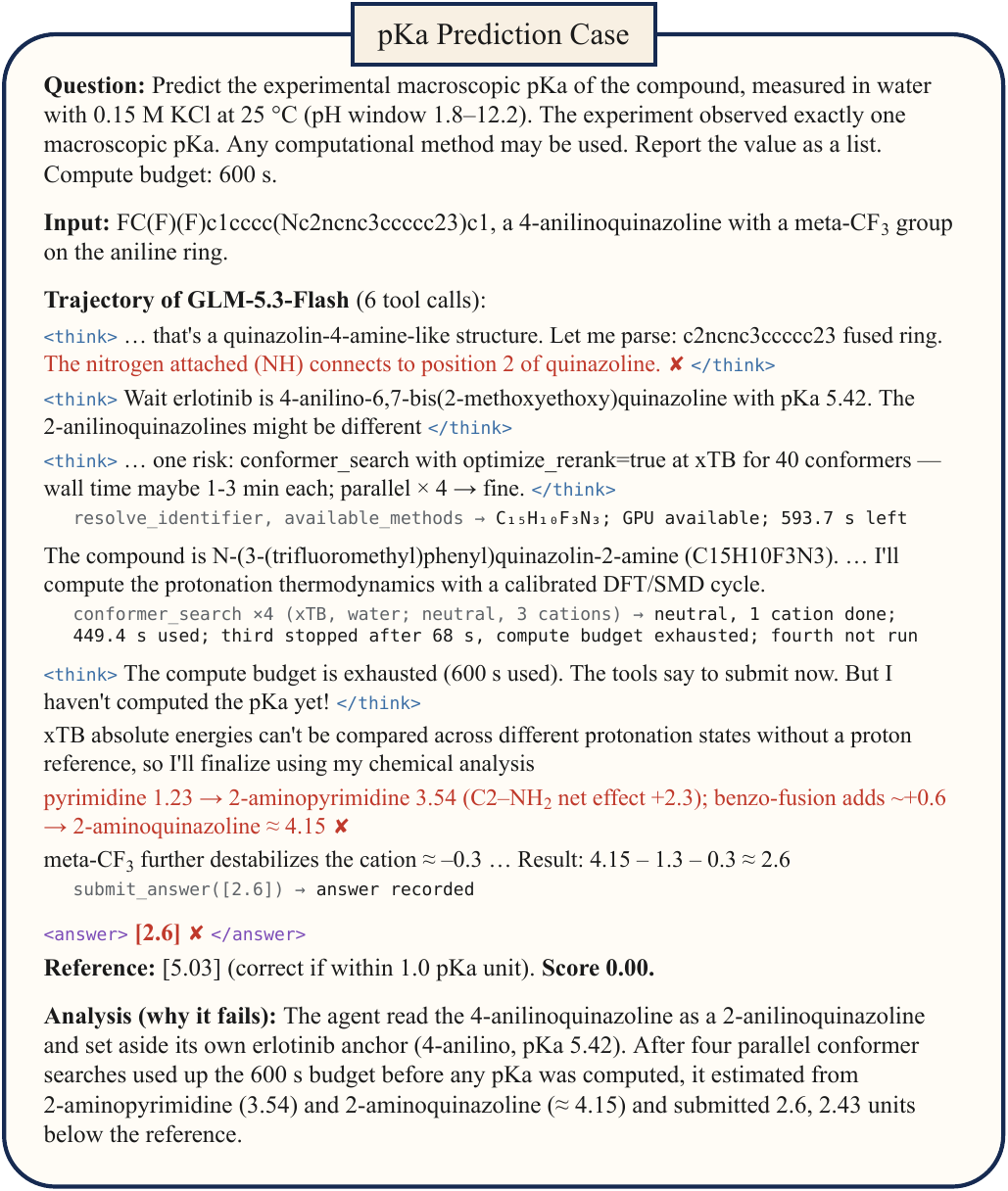}
  \caption{pKa prediction case (SIM-2, SIM2-SAMPLPKA-0029): GLM-5.3-Flash, incorrect.}
  \label{fig:case-sim2-samplpka-0029}
\end{figure*}

\begin{figure*}[p]
  \centering
  \includegraphics[width=\textwidth]{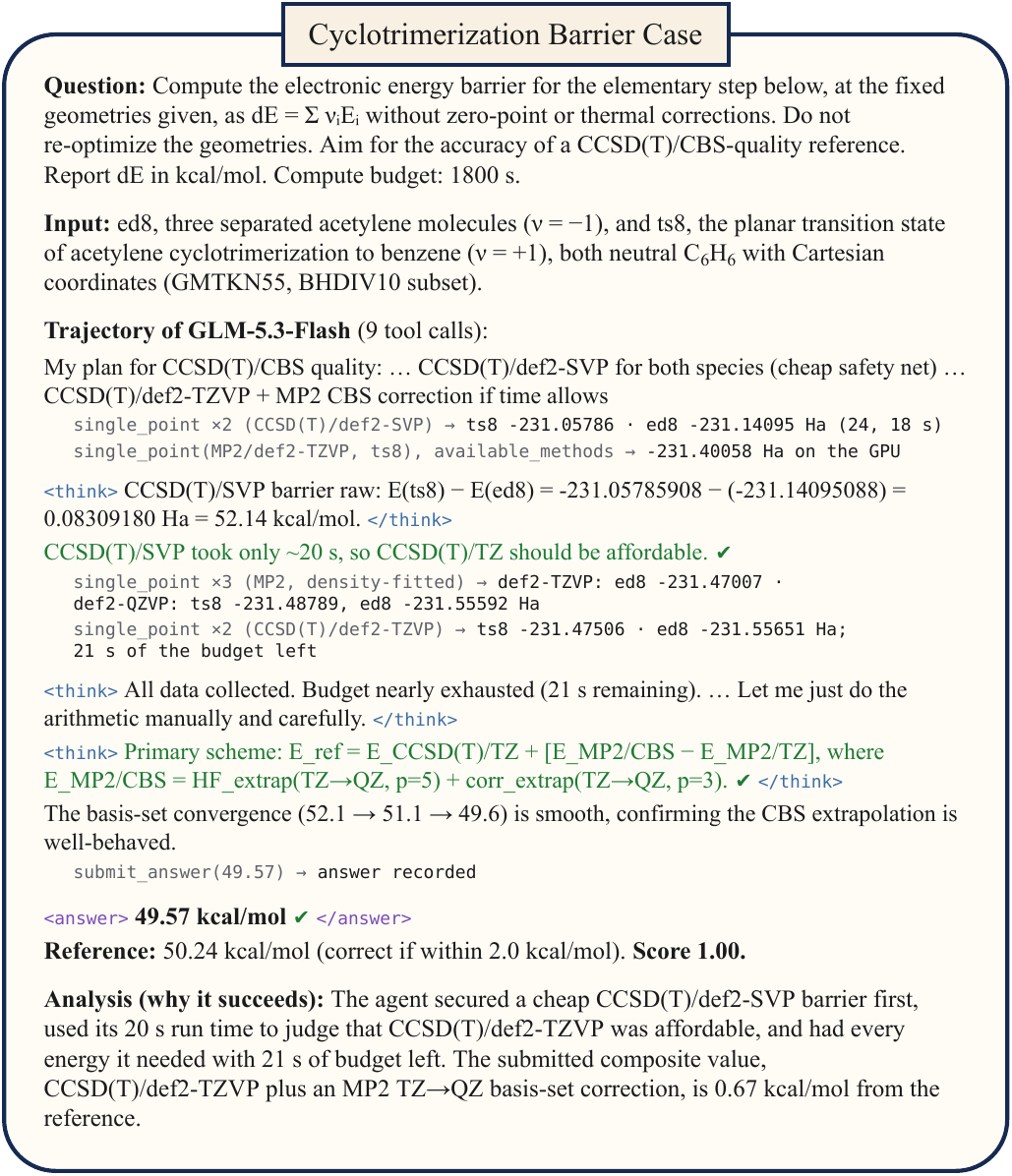}
  \caption{Cyclotrimerization barrier case (SIM-3, SIM3-GMTKN55-0010): GLM-5.3-Flash, correct.}
  \label{fig:case-sim3-gmtkn55-0010}
\end{figure*}

\begin{figure*}[p]
  \centering
  \includegraphics[width=\textwidth]{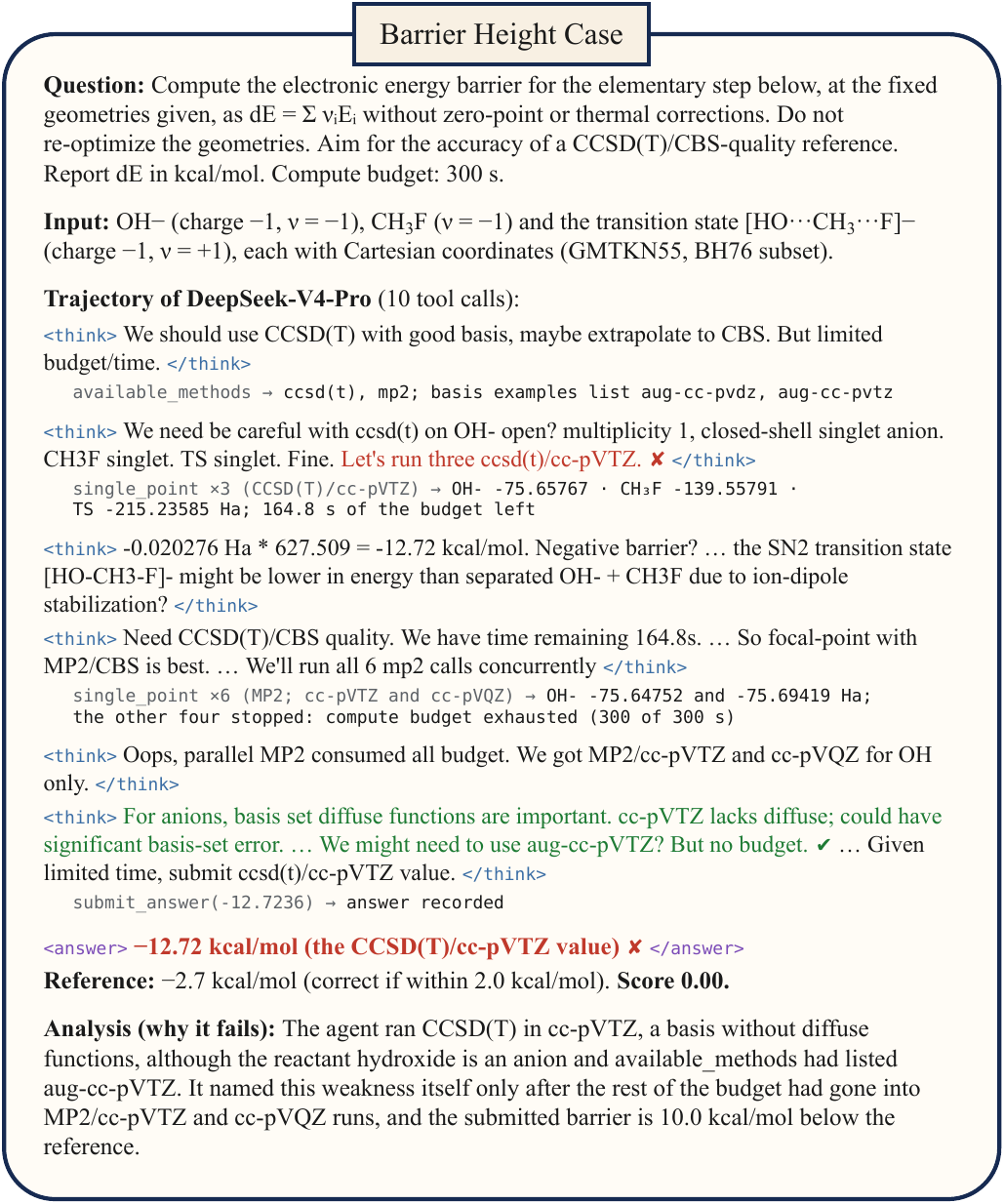}
  \caption{Barrier height case (SIM-3, SIM3-GMTKN55-0043): DeepSeek-V4-Pro, incorrect.}
  \label{fig:case-sim3-gmtkn55-0043}
\end{figure*}

\begin{figure*}[p]
  \centering
  \includegraphics[width=\textwidth]{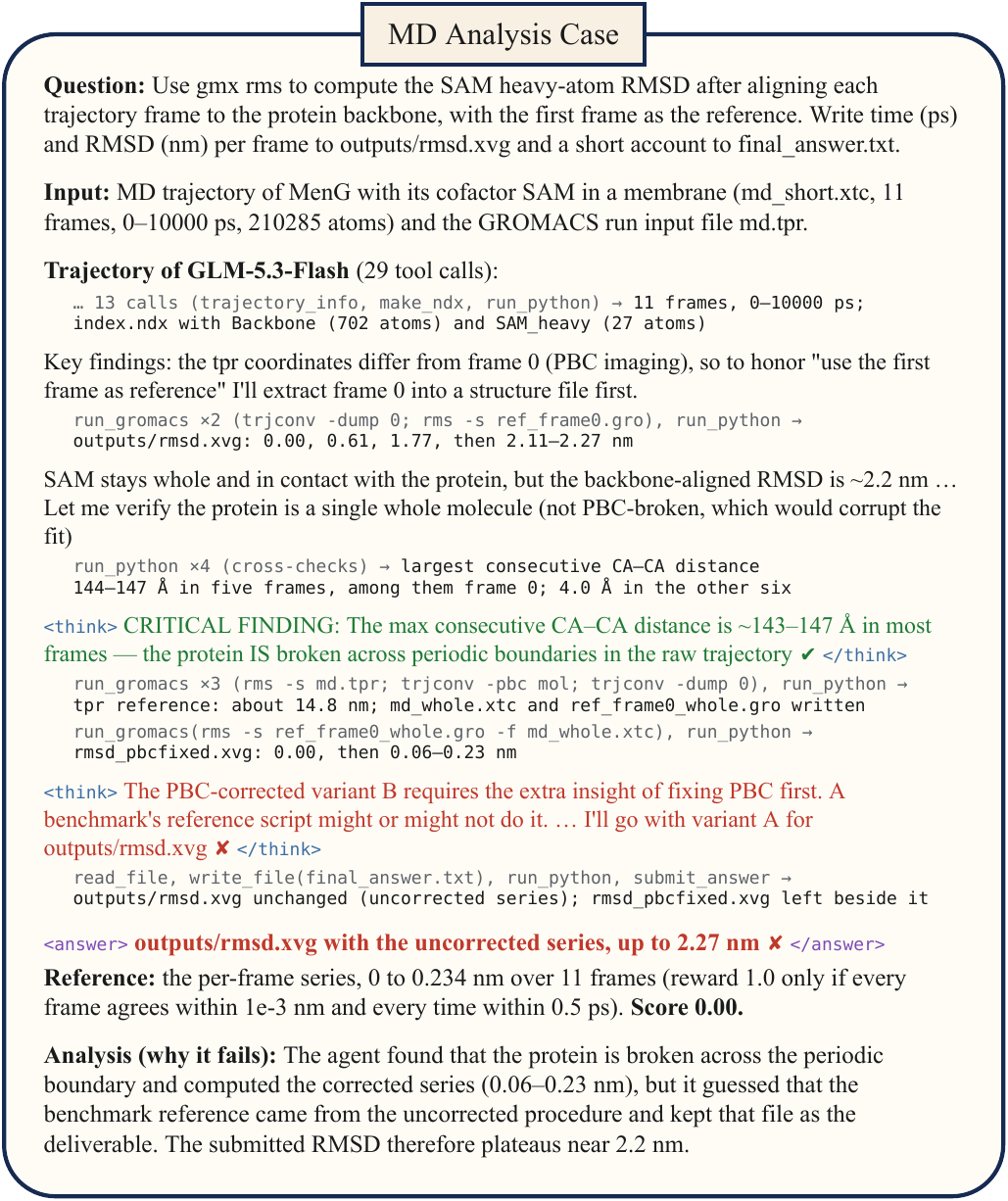}
  \caption{MD analysis case (SIM-4, SIM4-MDARENA-0012): GLM-5.3-Flash, score 0.}
  \label{fig:case-sim4-mdarena-0012}
\end{figure*}

\begin{figure*}[p]
  \centering
  \includegraphics[width=\textwidth]{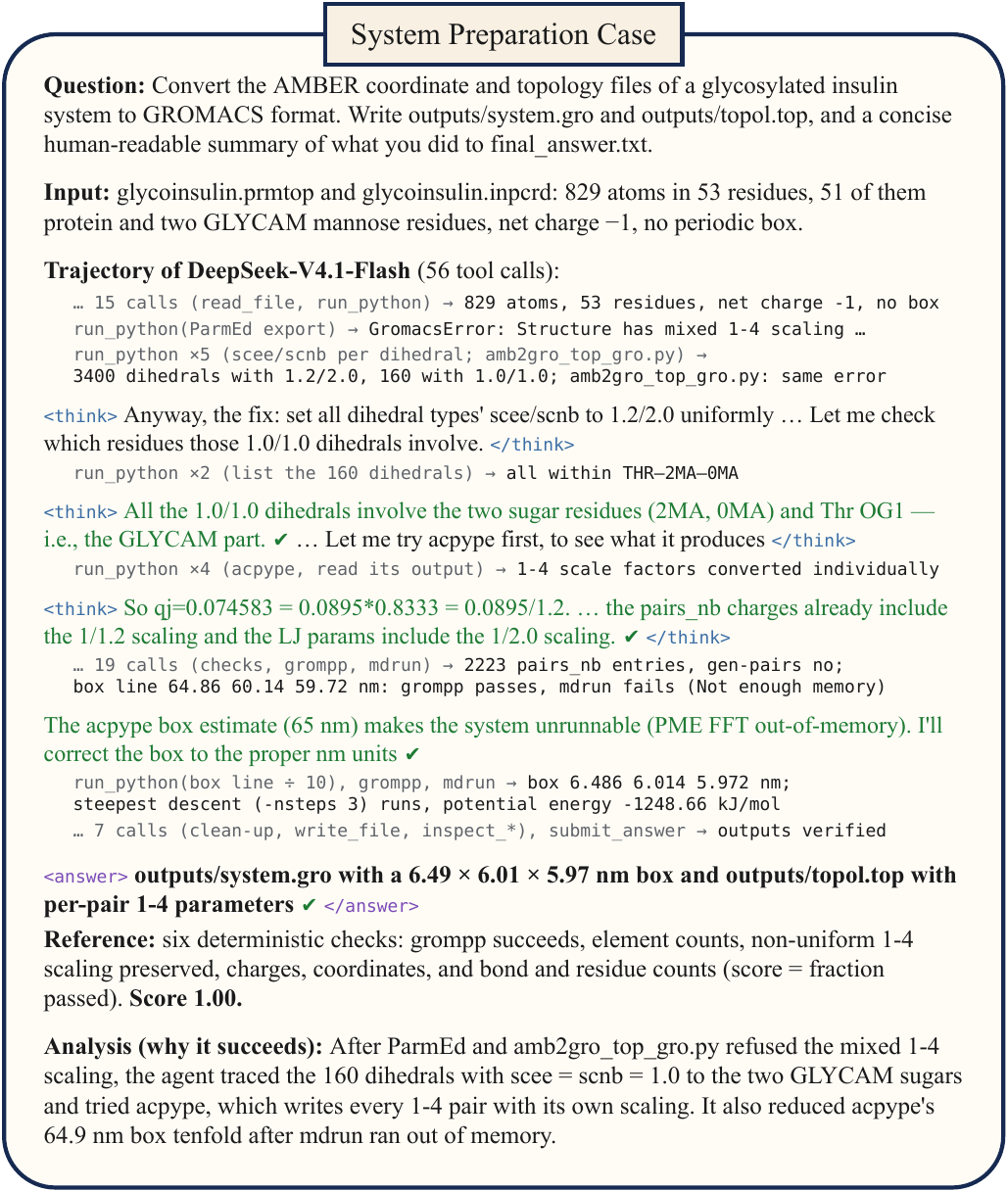}
  \caption{System preparation case (SIM-4, SIM4-MDARENA-0015): DeepSeek-V4.1-Flash, score 1.0.}
  \label{fig:case-sim4-mdarena-0015}
\end{figure*}

\begin{figure*}[p]
  \centering
  \includegraphics[width=\textwidth]{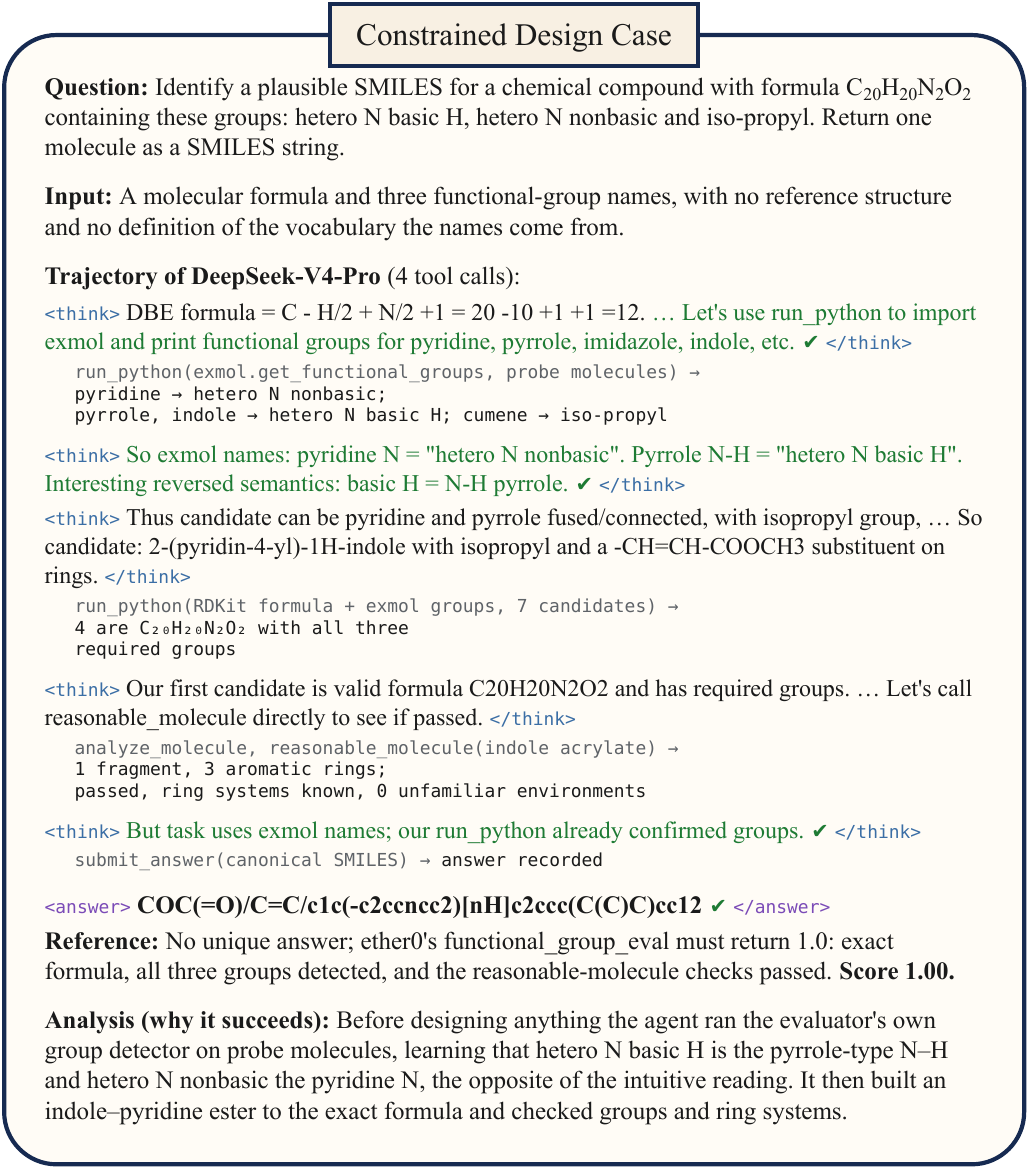}
  \caption{Constrained design case (DES-1, DES1-ETHER0-0010): DeepSeek-V4-Pro, correct.}
  \label{fig:case-des1-ether0-0010}
\end{figure*}

\begin{figure*}[p]
  \centering
  \includegraphics[width=\textwidth]{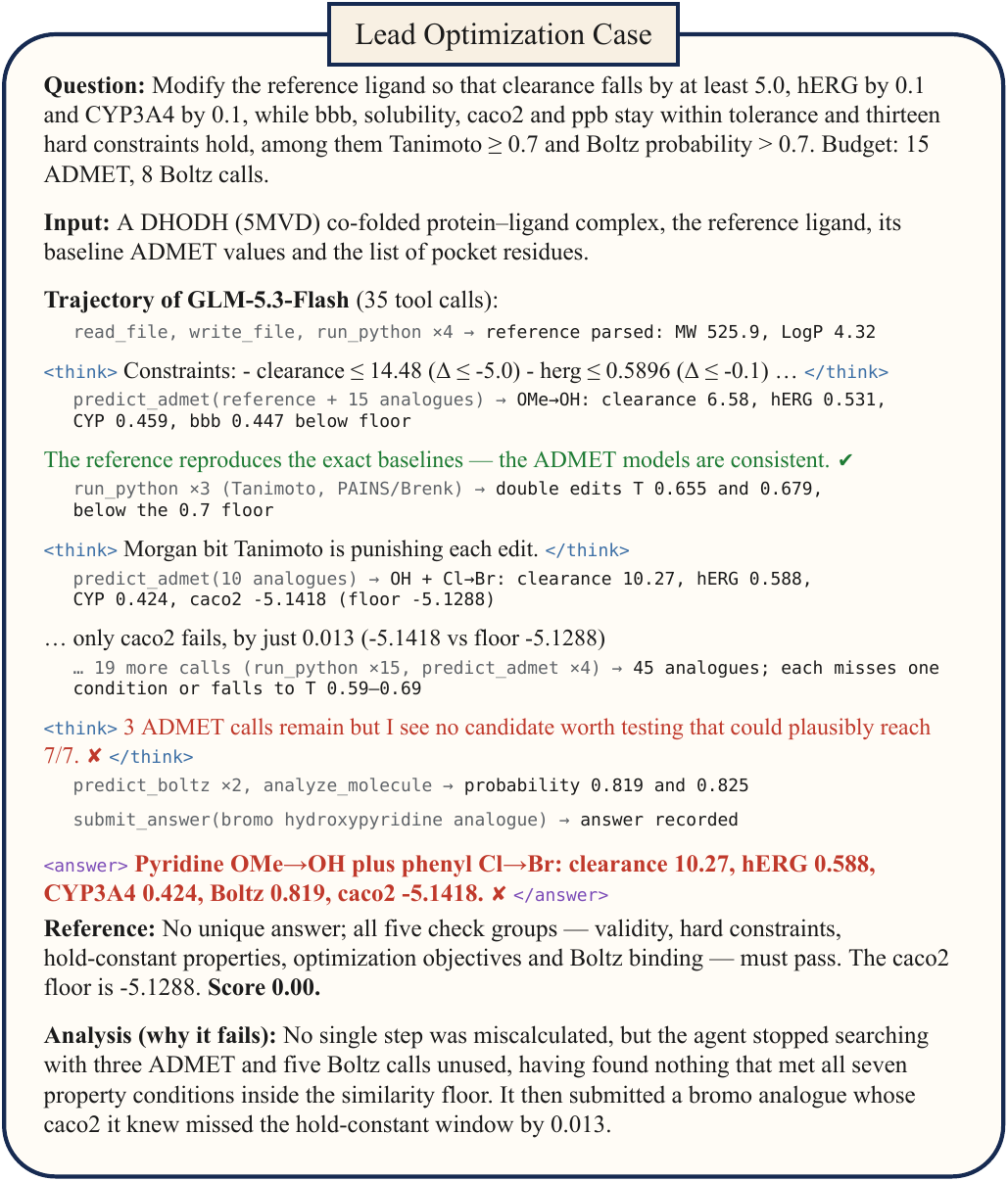}
  \caption{Lead optimization case (DES-1, DES1-SMDD-0089): GLM-5.3-Flash, incorrect.}
  \label{fig:case-des1-smdd-0089}
\end{figure*}

\begin{figure*}[p]
  \centering
  \includegraphics[width=\textwidth]{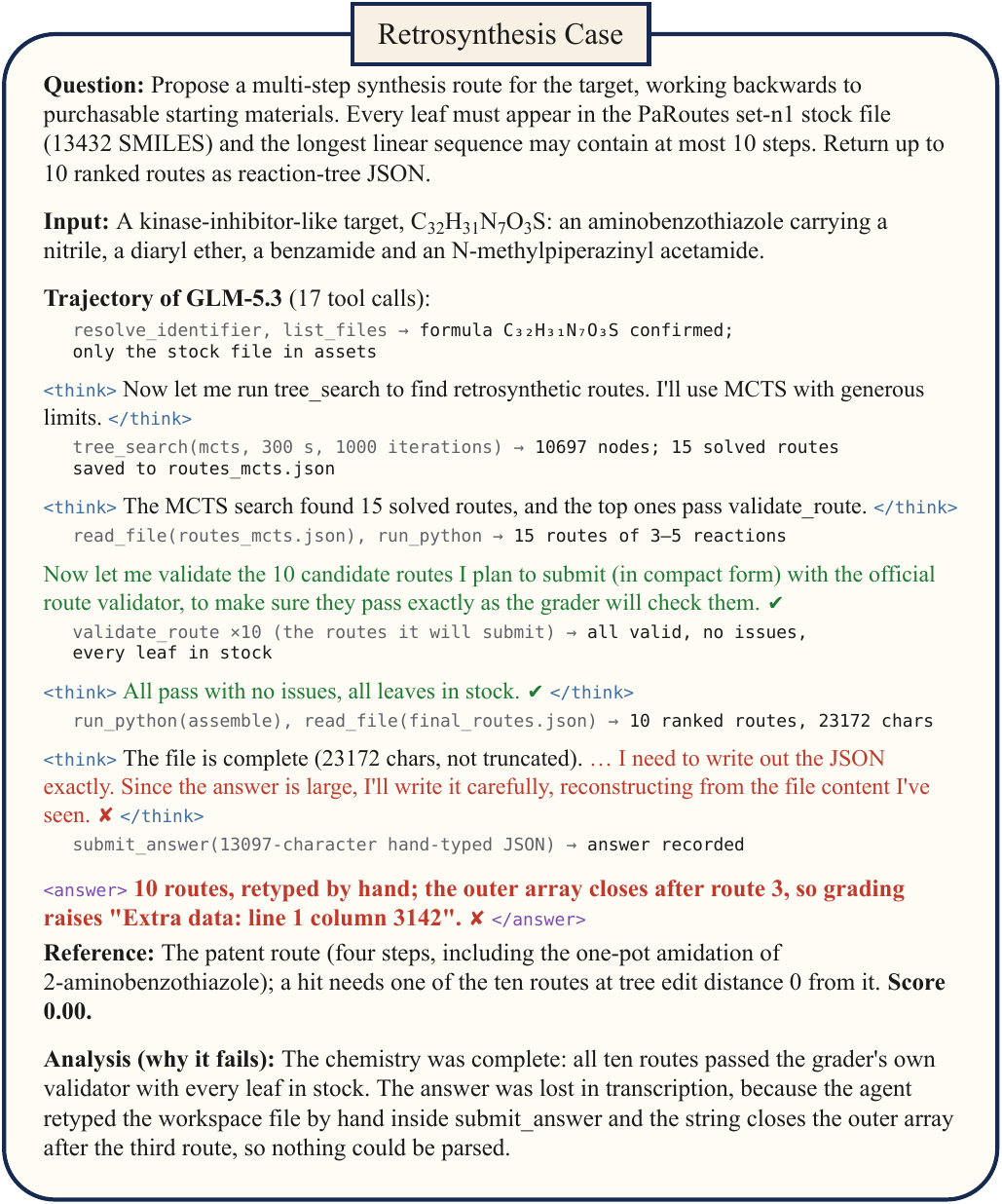}
  \caption{Retrosynthesis case (SYN-1, SYN1-PAROUTES-0050): GLM-5.3, incorrect.}
  \label{fig:case-syn1-paroutes-0050}
\end{figure*}

\begin{figure*}[p]
  \centering
  \includegraphics[width=\textwidth]{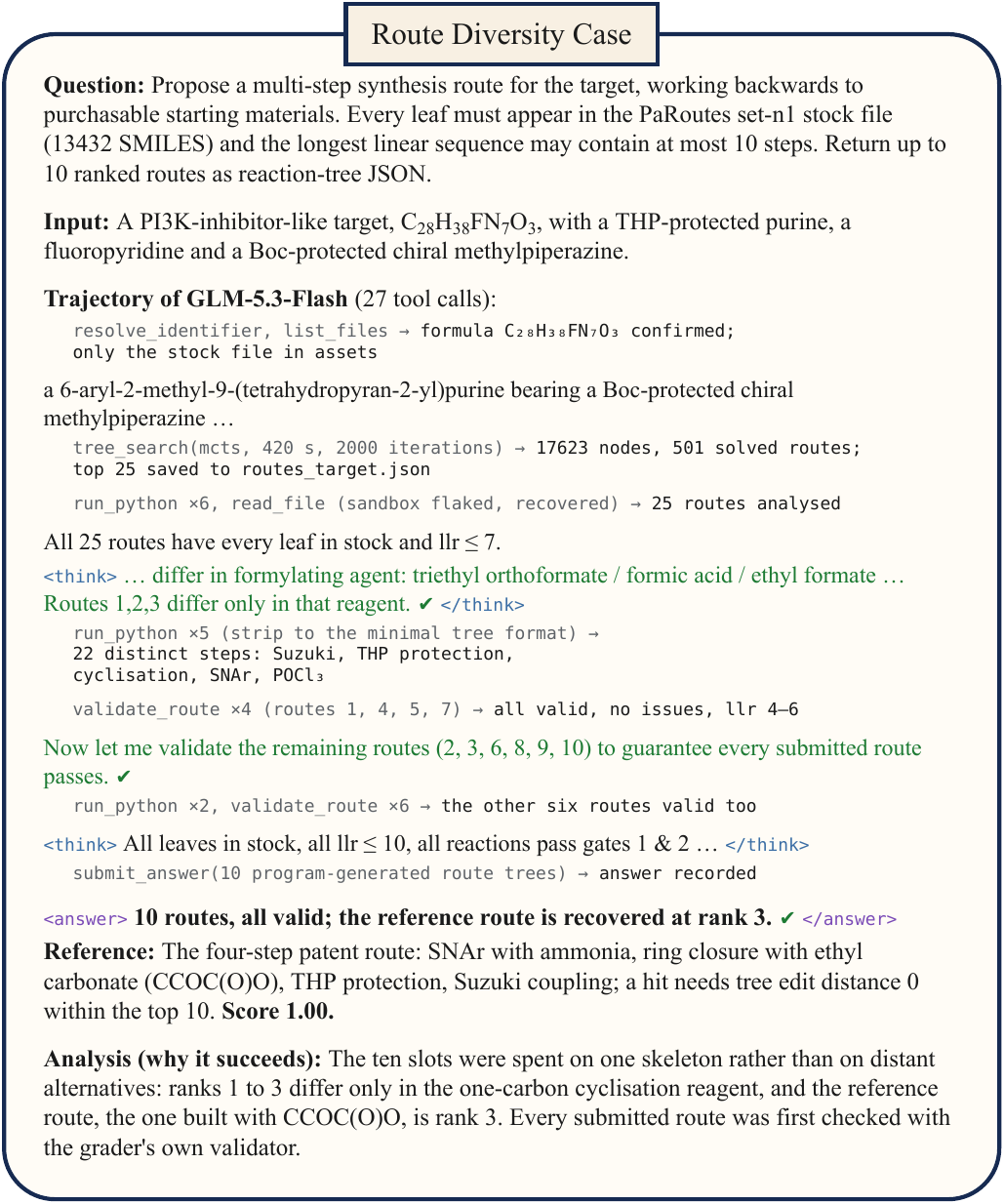}
  \caption{Route diversity case (SYN-1, SYN1-PAROUTES-0057): GLM-5.3-Flash, correct.}
  \label{fig:case-syn1-paroutes-0057}
\end{figure*}

\begin{figure*}[p]
  \centering
  \includegraphics[width=\textwidth]{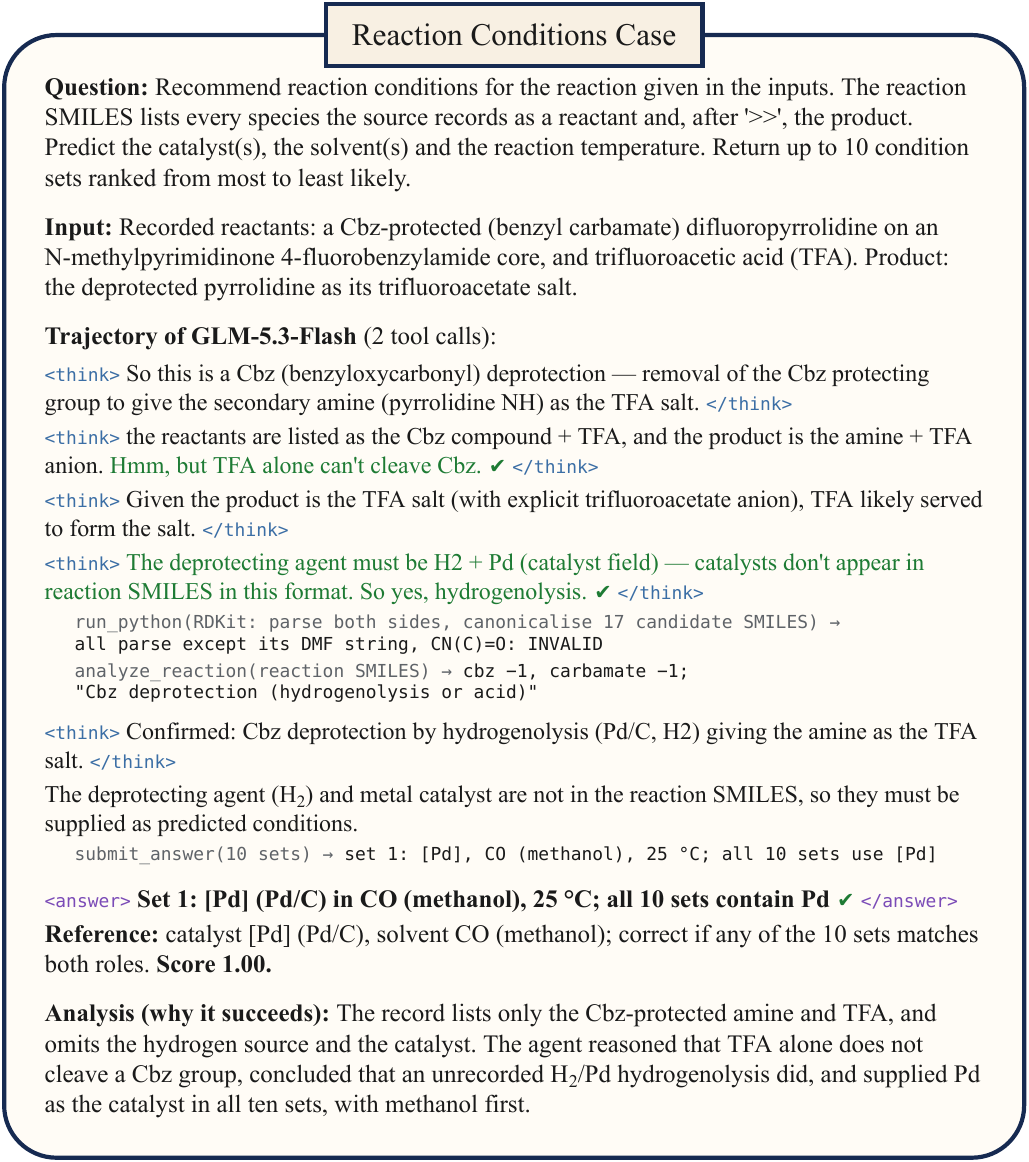}
  \caption{Reaction conditions case (SYN-2, SYN2-ORDUSPTO-0118): GLM-5.3-Flash, correct.}
  \label{fig:case-syn2-orduspto-0118}
\end{figure*}

\begin{figure*}[p]
  \centering
  \includegraphics[width=\textwidth]{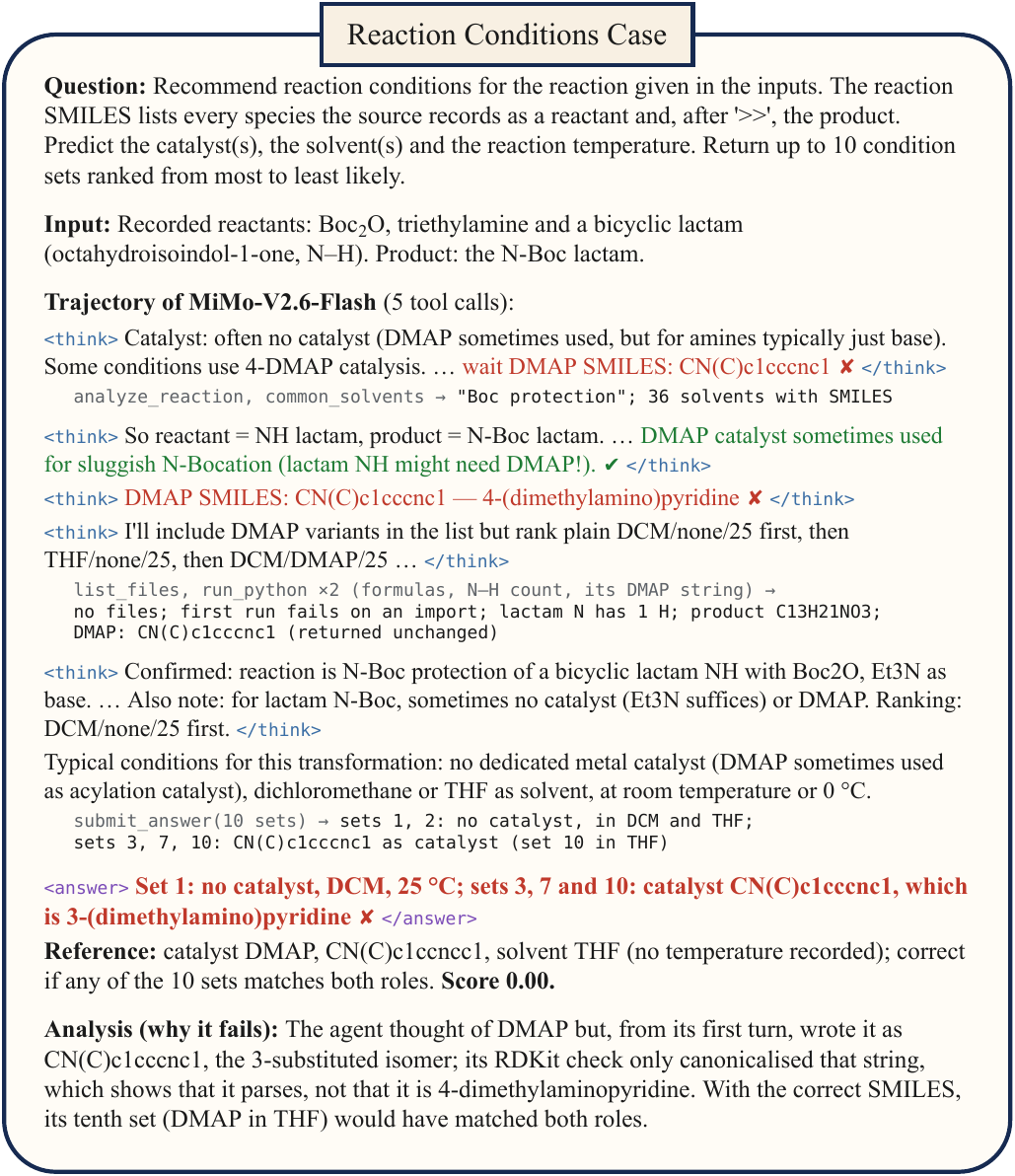}
  \caption{Reaction conditions case (SYN-2, SYN2-ORDUSPTO-0110): MiMo-V2.6-Flash, incorrect.}
  \label{fig:case-syn2-orduspto-0110}
\end{figure*}

\begin{figure*}[p]
  \centering
  \includegraphics[width=\textwidth]{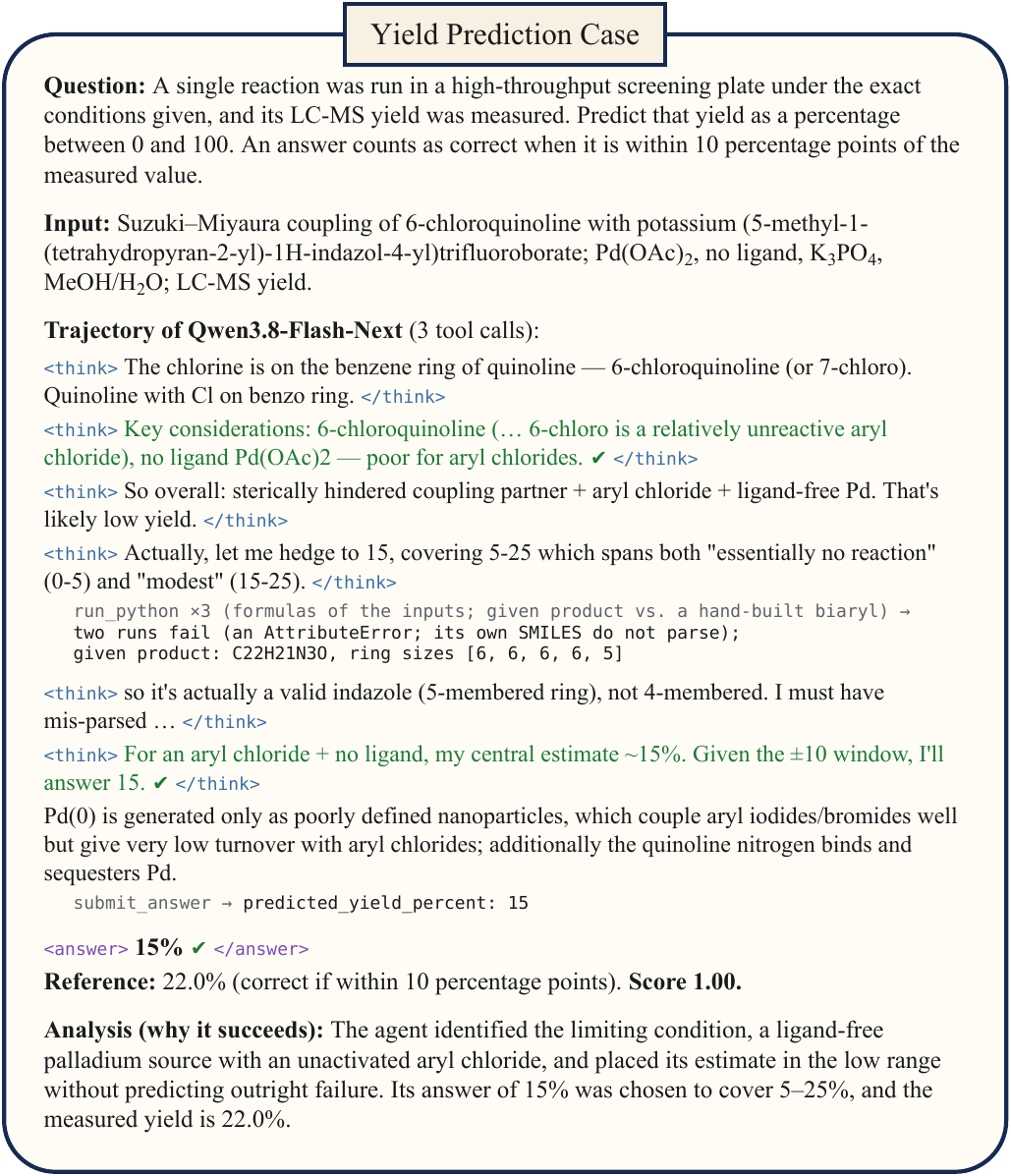}
  \caption{Yield prediction case (SYN-3, SYN3-HTEYIELD-0015): Qwen3.8-Flash-Next, correct.}
  \label{fig:case-syn3-hteyield-0015}
\end{figure*}

\begin{figure*}[p]
  \centering
  \includegraphics[width=\textwidth]{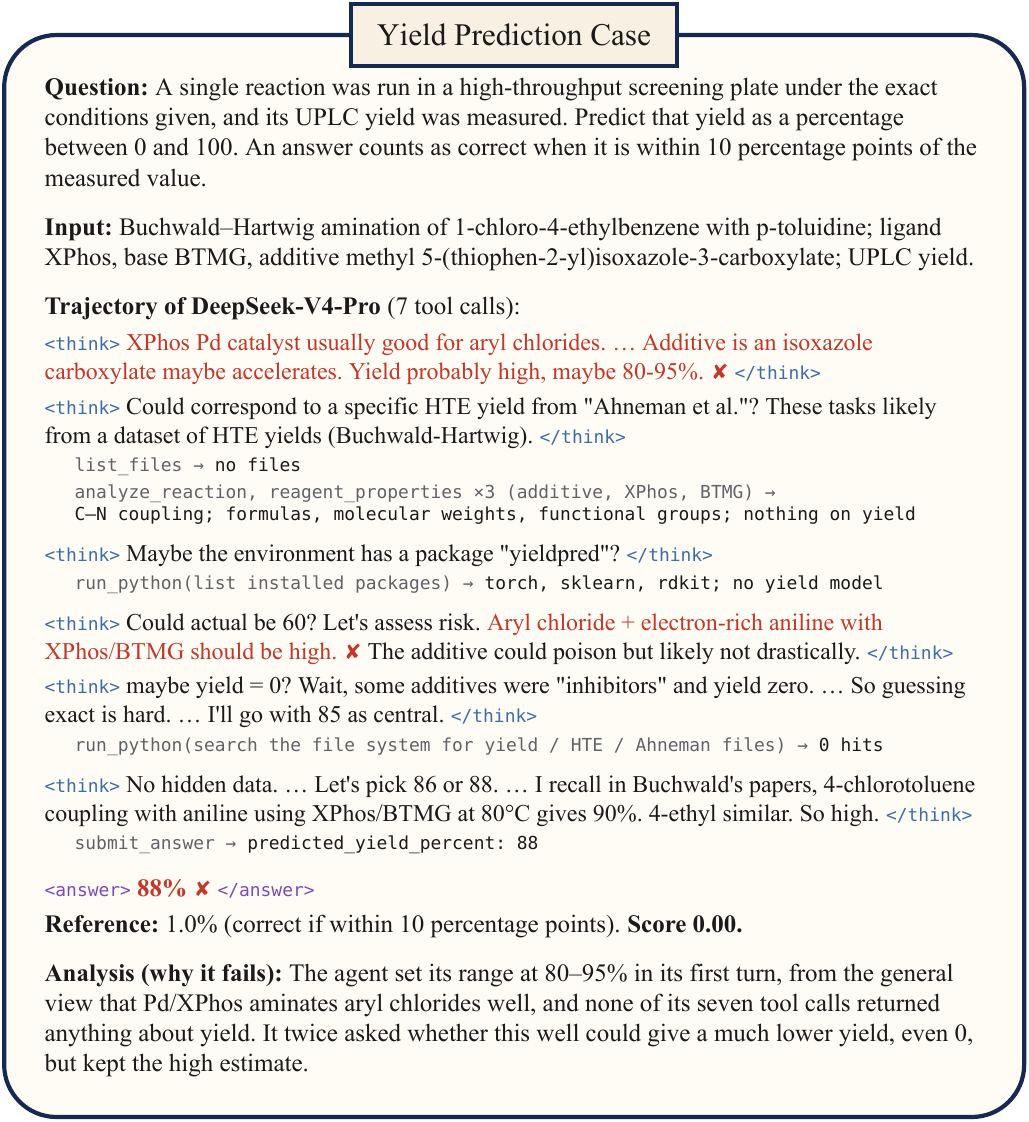}
  \caption{Yield prediction case (SYN-3, SYN3-HTEYIELD-0001): DeepSeek-V4-Pro, incorrect.}
  \label{fig:case-syn3-hteyield-0001}
\end{figure*}

\begin{figure*}[p]
  \centering
  \includegraphics[width=\textwidth]{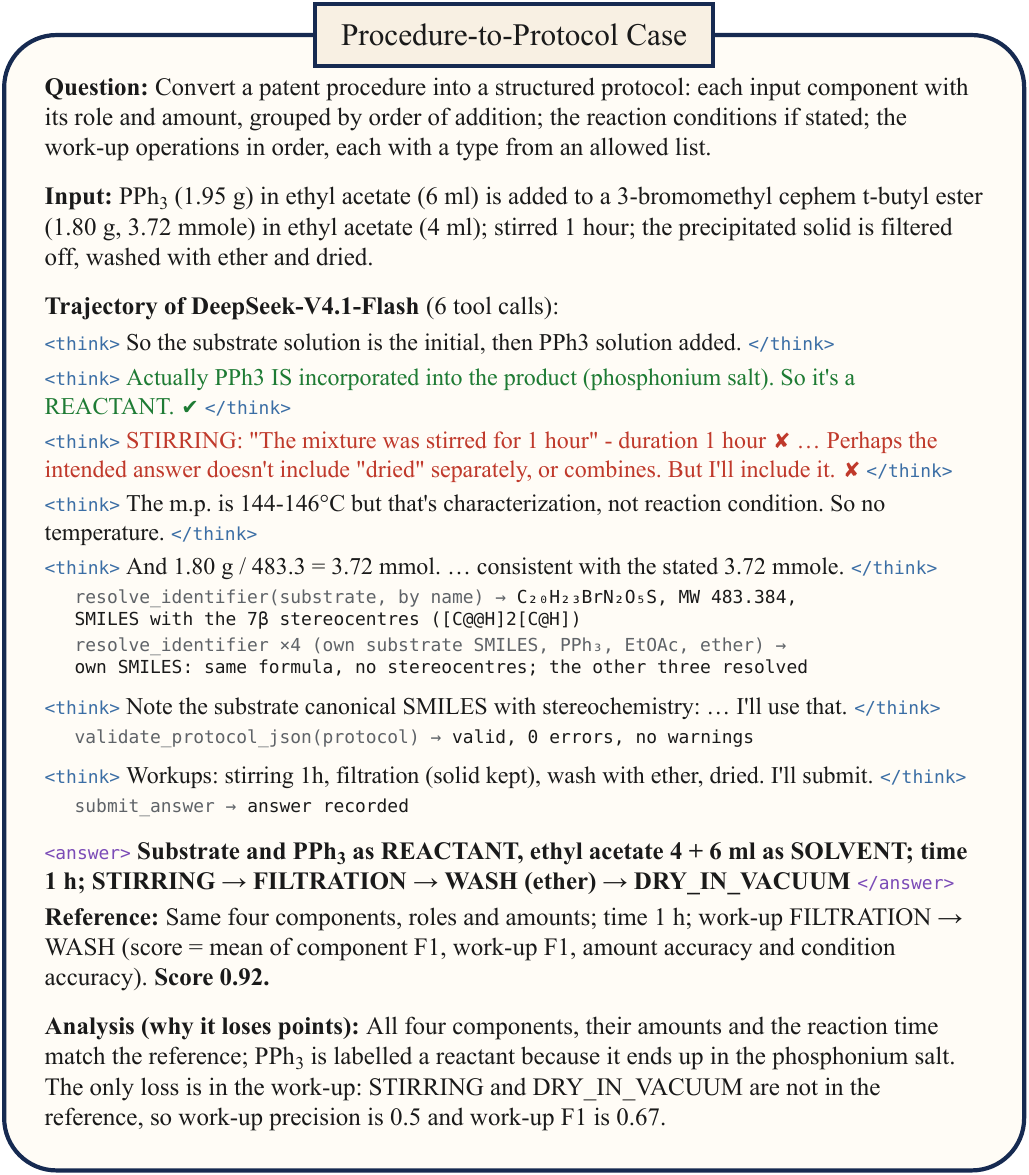}
  \caption{Procedure-to-protocol case (EXE-1, EXE1-ORDUSPTO-0015): DeepSeek-V4.1-Flash, score 0.92.}
  \label{fig:case-exe1-orduspto-0015}
\end{figure*}

\begin{figure*}[p]
  \centering
  \includegraphics[width=\textwidth]{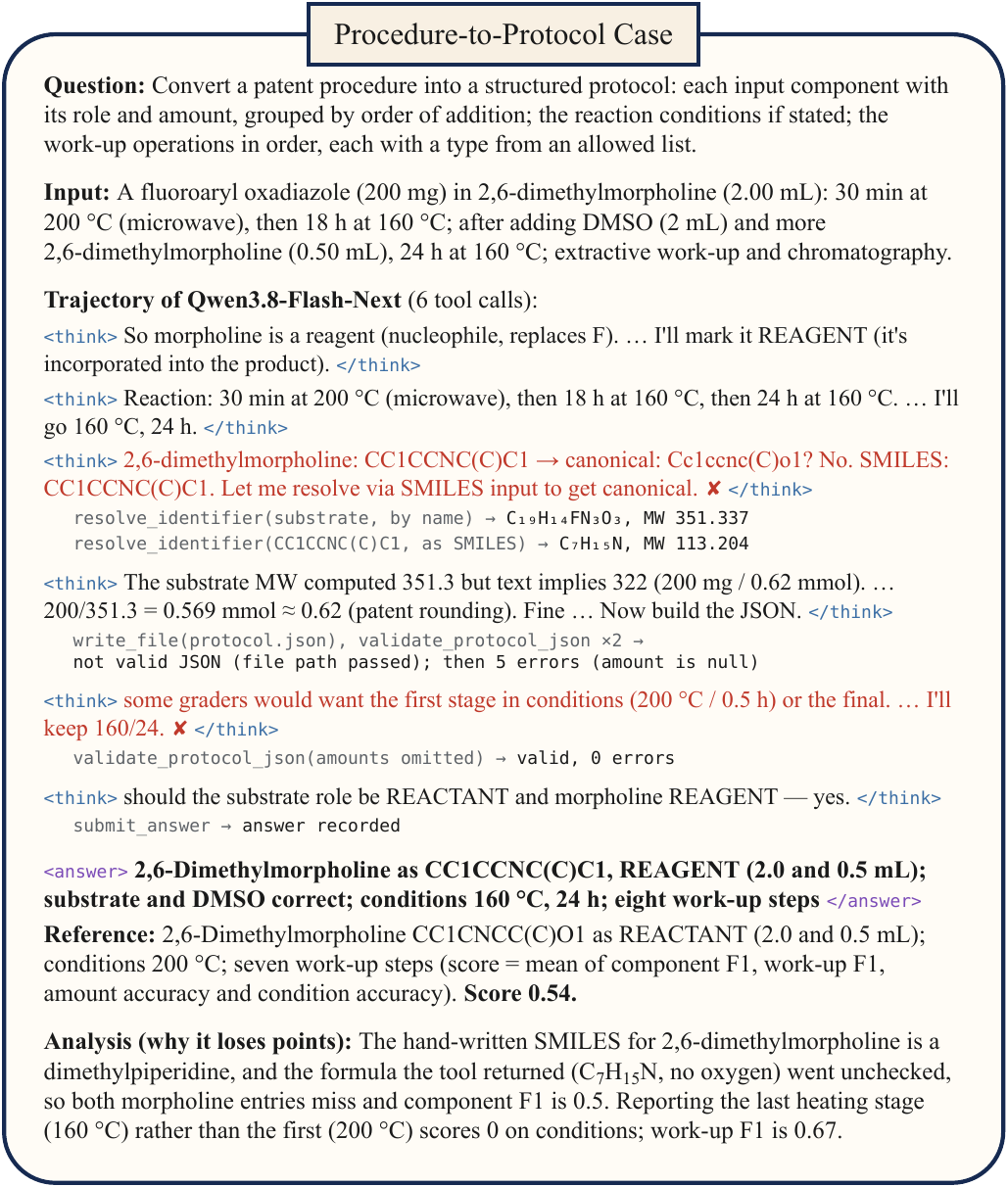}
  \caption{Procedure-to-protocol case (EXE-1, EXE1-ORDUSPTO-0099): Qwen3.8-Flash-Next, score 0.54.}
  \label{fig:case-exe1-orduspto-0099}
\end{figure*}

\begin{figure*}[p]
  \centering
  \includegraphics[width=\textwidth]{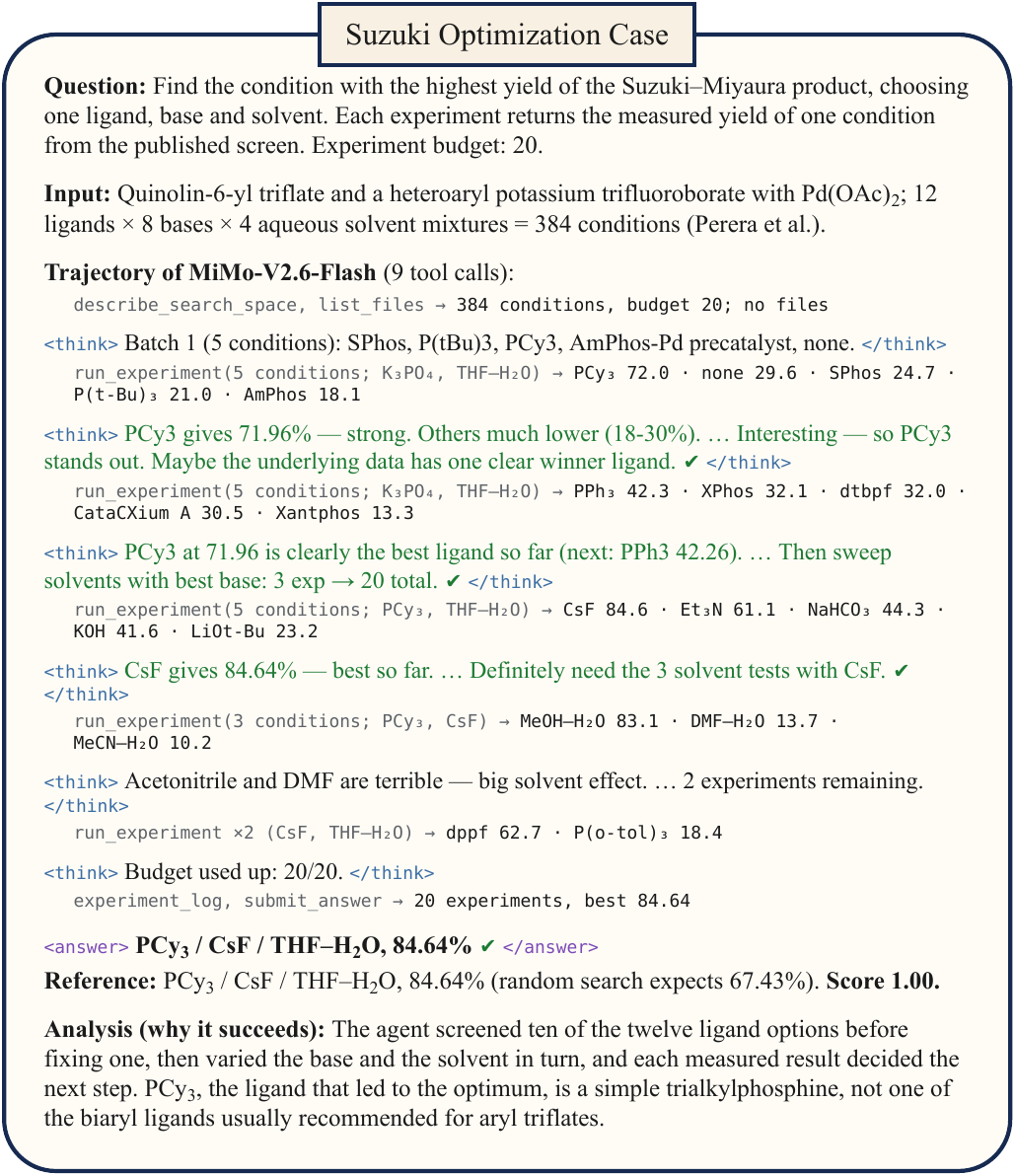}
  \caption{Suzuki optimization case (OPT-1, OPT1-PERERASUZUKI-0002): MiMo-V2.6-Flash, score 1.00.}
  \label{fig:case-opt1-pererasuzuki-0002}
\end{figure*}

\begin{figure*}[p]
  \centering
  \includegraphics[width=\textwidth]{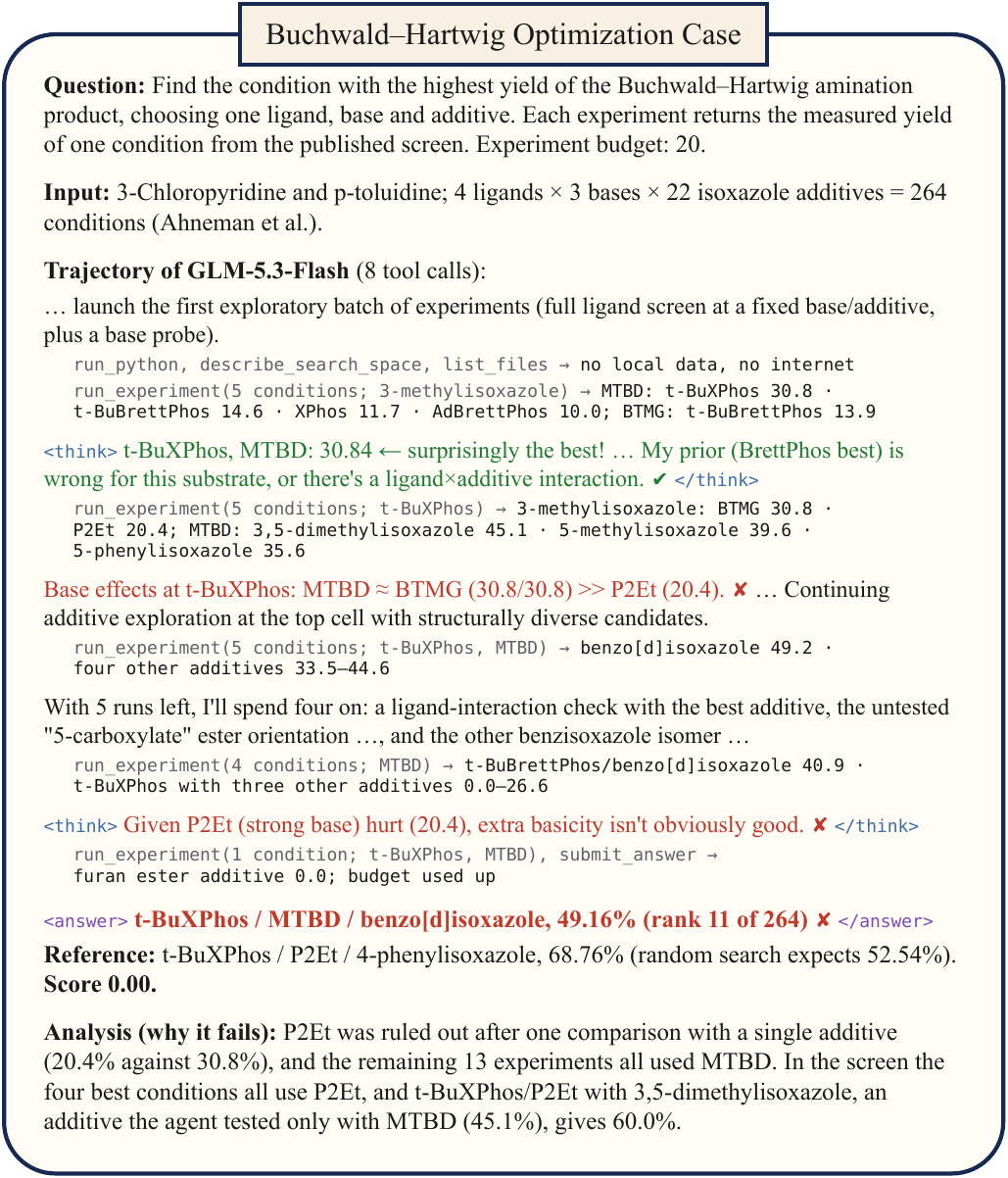}
  \caption{Buchwald--Hartwig optimization case (OPT-1, OPT1-AHNEMANBH-0026): GLM-5.3-Flash, score 0.00.}
  \label{fig:case-opt1-ahnemanbh-0026}
\end{figure*}

\begin{figure*}[p]
  \centering
  \includegraphics[width=\textwidth]{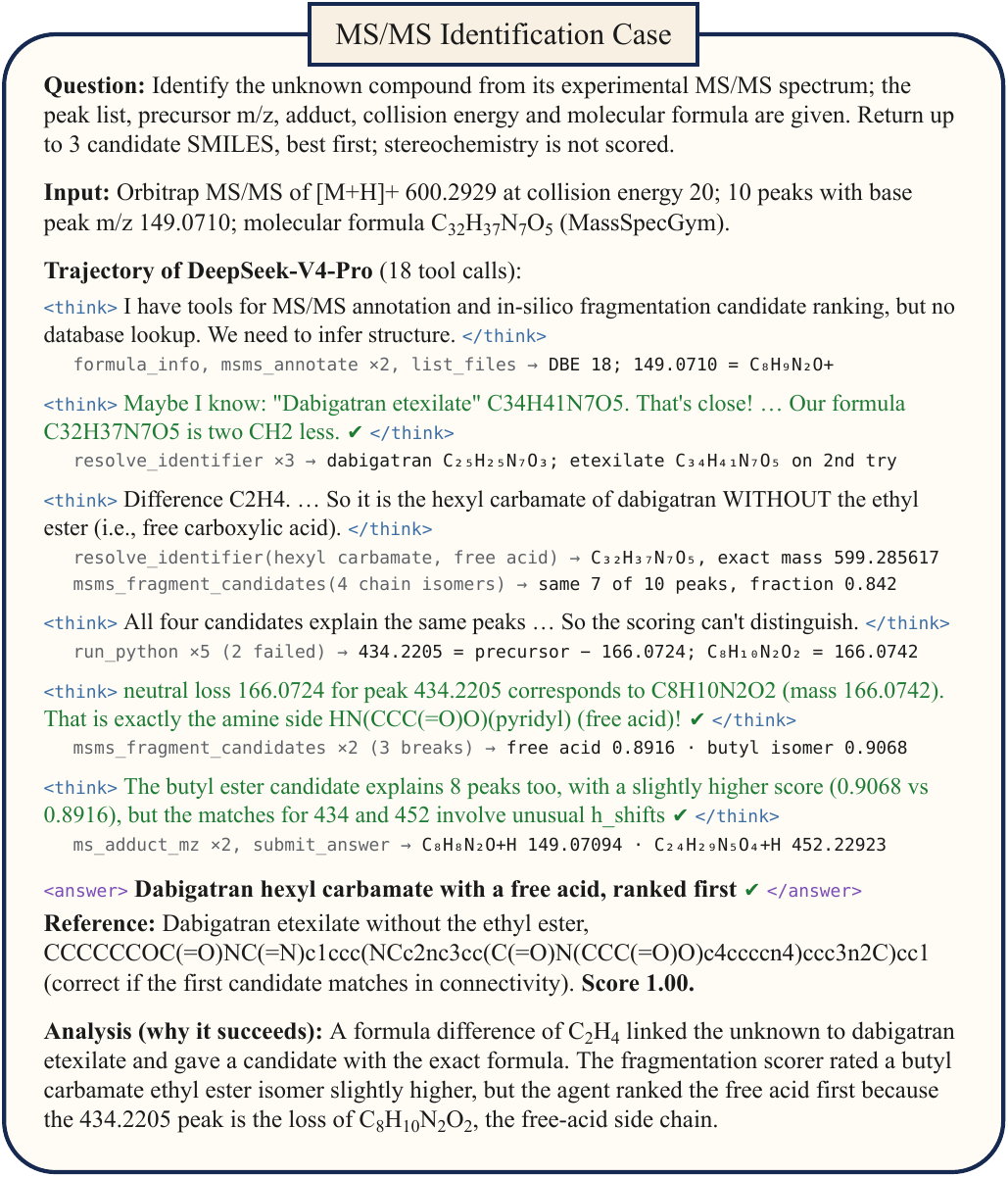}
  \caption{MS/MS identification case (ANA-1, ANA1-MSGYM-0027): DeepSeek-V4-Pro, correct.}
  \label{fig:case-ana1-msgym-0027}
\end{figure*}

\begin{figure*}[p]
  \centering
  \includegraphics[width=\textwidth]{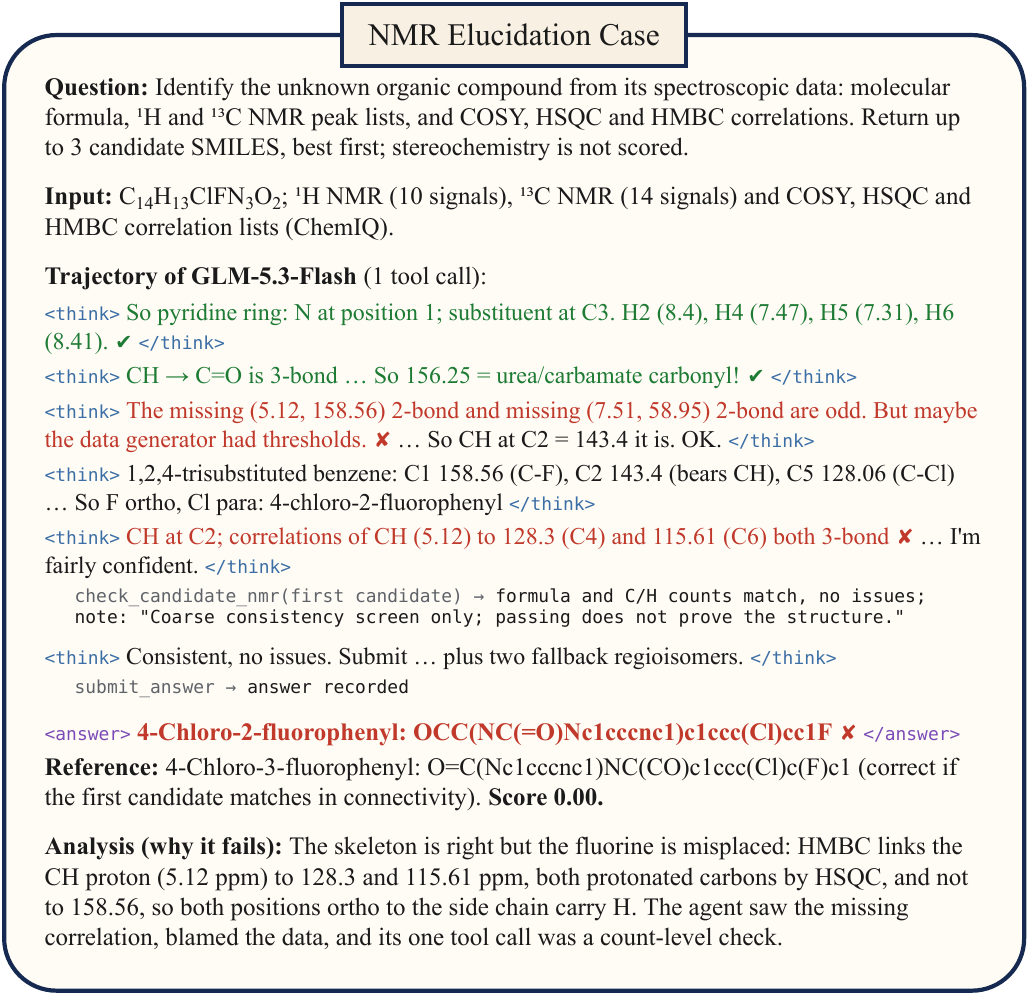}
  \caption{NMR elucidation case (ANA-1, ANA1-CHEMIQ-0032): GLM-5.3-Flash, incorrect.}
  \label{fig:case-ana1-chemiq-0032}
\end{figure*}

\begin{figure*}[p]
  \centering
  \includegraphics[width=\textwidth]{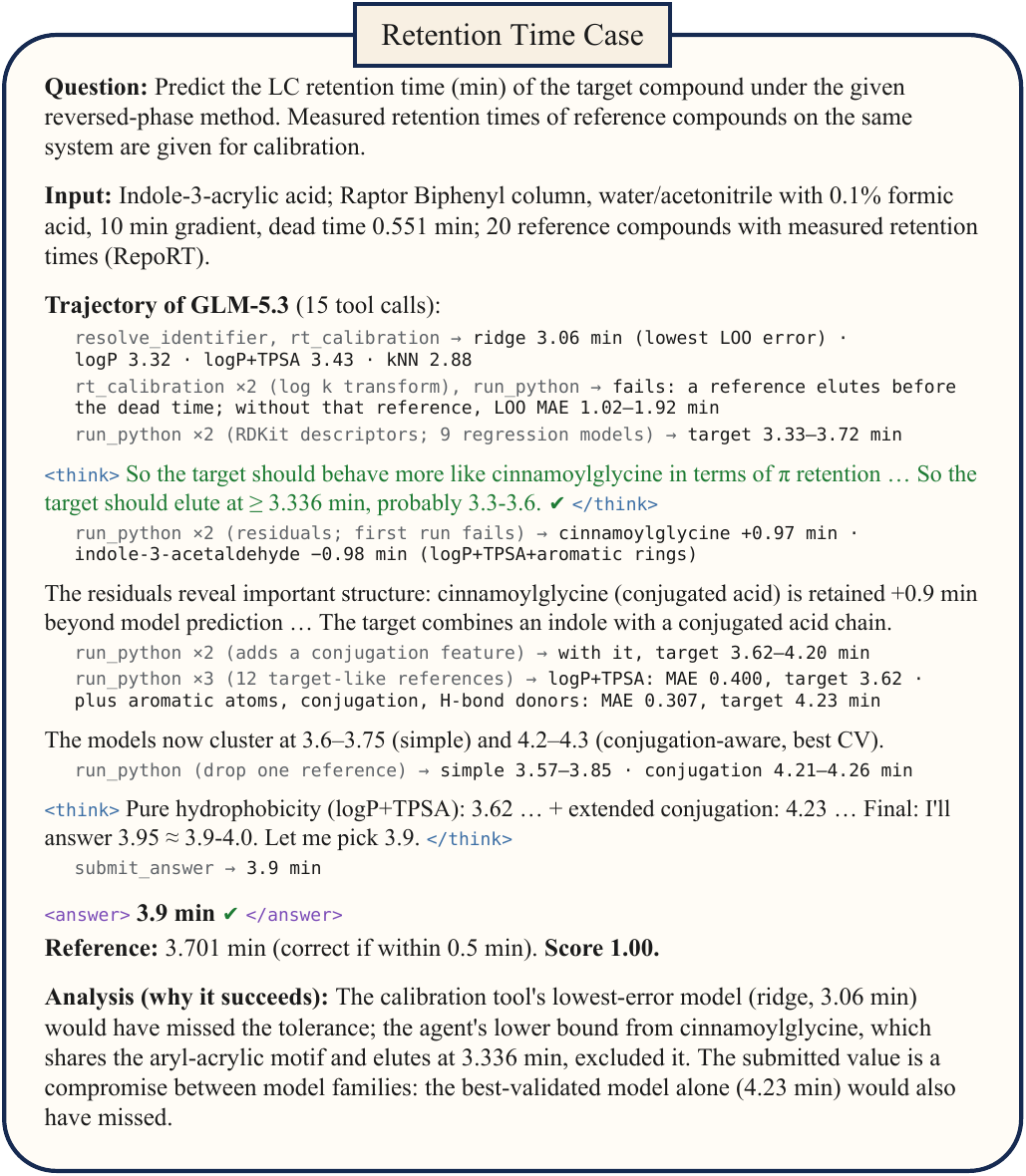}
  \caption{Retention time case (ANA-2, ANA2-REPORT-0019): GLM-5.3, correct.}
  \label{fig:case-ana2-report-0019}
\end{figure*}

\begin{figure*}[p]
  \centering
  \includegraphics[width=\textwidth]{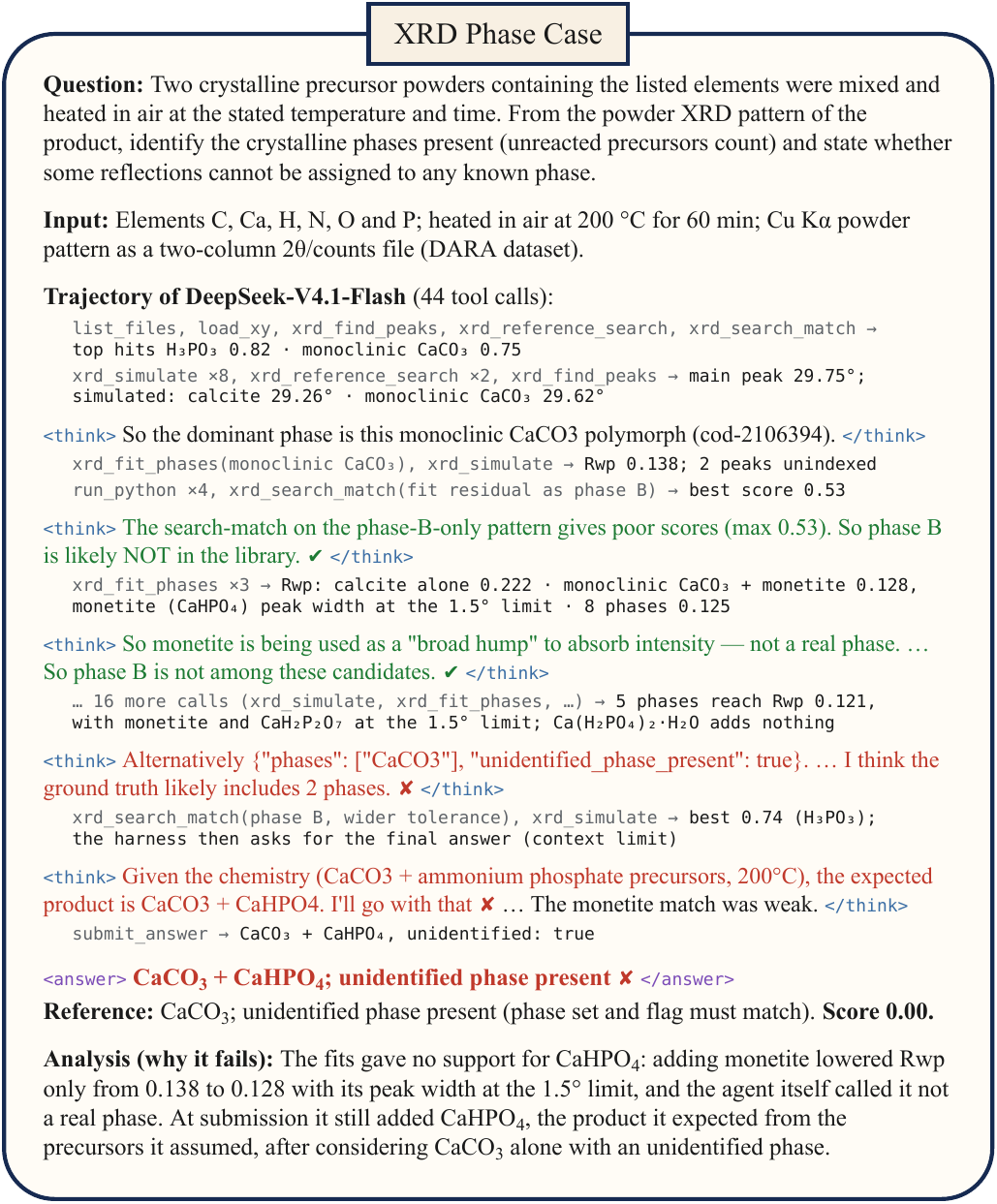}
  \caption{XRD phase case (ANA-2, ANA2-DARA-0027): DeepSeek-V4.1-Flash, incorrect.}
  \label{fig:case-ana2-dara-0027}
\end{figure*}

Figures~\ref{fig:case-sim1-gmtkn55-0096}--\ref{fig:case-ana2-dara-0027} show two cases for each of the 12 task families. Each case gives the task, the trajectory of one model, its output against the ground truth, and an analysis of why it succeeded or failed.

